\documentclass[a4paper]{sig-alternate}
\usepackage{comment}
\usepackage{algorithm}
\usepackage{algorithmic}
\usepackage{subfigure}
\usepackage{graphicx}
\usepackage{url}
 \usepackage[centerlast]{caption2}
\usepackage{subfigure}
\usepackage[normalem]{ulem}
\usepackage{color}

\begin{document}

\title{Low Rank Representation on Riemannian Manifold of Square Root Densities}
%
%
%
%
%

\numberofauthors{4} 
%

\author{
%
%
\alignauthor
Yifan Fu and Junbin Gao
\\
       \affaddr{School of Computing and Mathematics}\\
       \affaddr{ Charles Sturt University}\\
       \affaddr{Bathurst, NSW 2795, Australia}\\
       \email{\{yfu,jbgao\}@csu.edu.au}
\alignauthor Xia Hong 
\\
       \affaddr{School of Systems Engineering}\\
       \affaddr{University of Reading}\\
       \affaddr{Reading, RG6 6AY, UK}\\
       \email{x.hong@reading.ac.uk}
\and  
\alignauthor David Tien
\\
       \affaddr{School of Computing and Mathematics}\\
       \affaddr{Charles Sturt University}\\
       \affaddr{Bathurst, NSW 2795, Australia}\\
       \email{dtien@csu.edu.au}}
\maketitle
\begin{abstract}
In this paper, we present a novel low rank representation (LRR) algorithm for data lying on the manifold of square root densities. Unlike traditional
LRR methods which rely on the assumption that the data points are vectors in the Euclidean space, our new algorithm is designed to incorporate the intrinsic geometric structure  and geodesic distance of the manifold. Experiments on several computer vision  datasets showcase its noise robustness and superior performance on classification and subspace clustering compared to other state-of-the-art approaches.

\end{abstract}




\section{Introduction}
 Dictionary learning has been proven very effective to find sparse representation for high dimensional data, widely used in many machine learning applications, such as classification \cite{KongWang2012}, recognition \cite{LiuYangGaoYinChen2014} and image restoration \cite{BaoCaiJi2013}. Under this model, each data point can be recovered/ or represented by using a linear combination of a small number of atoms under a dictionary. The underlying linear process requires that  both data points and the atoms are from a linear space embedded in an Euclidean space and the reconstruction error is measured using $l_2$-norm. As learning a dictionary is quite time consuming, the data point themselves are generally used as a dictionary based on the so-called self-representative principle \cite{ElhamifarVidal2013}. Given a collection of  data points $X=\{\mathbf{x}_1,\cdots,\mathbf{x}_n\}$ consisting of $n$ ${m+1}$-dimensional column vectors $x_i$ ($i=1,2,\ldots, n$), the self-representative coding seeks a joint sparse representation of $X$ using data points themselves as the dictionary, which can be formulated as,
\begin{equation}\label{lrrg}
\min_{W}\|X-XW\|^2+\lambda \|W\|_{q}
\end{equation}
where $W =(\mathbf w_1, \ldots, \mathbf w_n)$ is the sparse representation matrix  under the dictionary $X$, each column $\mathbf w_i$ ($i=1, \ldots, n$) is the corresponding  representation of $x_i$,  $\|\cdot\|_{q}$ is the sparsity regularizer term for the new representation $W$ and $\lambda >0$ is a penalty parameter to balance the sparsity regularizer term and the reconstruction error.
In general, the $l1$ norm $\|W\|_{1}$  (i.e. the sum of the absolute values of all the element in a matrix) is used in favor of independent sparse representation for each data point, while the nuclear norm $\|W\|_{\star}$ (i.e. the sum of all the singular values of a matrix) is employed to holistically reveal the latent sparse property embedded in the whole data set.

However, in the context of machine learning and vision applications, feature data acquired actually satisfy extra constraints which make them so-called manifold-valued. Take diffusion weighted magnetic resonance imaging (MRI) \cite{Jones2011} as an example, this non-invasive imaging technique helps explore the complex micro-structure of fibrous tissues through sensing the Brownian motion of water molecules.
Water diffusion is fully characterized by the diffusion probability density function called the diffusion propagator (DP) \cite{Callaghan1994}, which can be represented in the form of the square root density functions. More specifically, for a probability density function $p$ and its continuous square root $\psi=\sqrt p$, we have
\begin{equation}\label{sq}
\int_S \psi^2ds=1
\end{equation}

$\psi$ in Eq. \eqref{sq} can be identified as  a point on the unit sphere in a Hilbert space \cite{SrivastavaJermynJoshi2007} by expanding it using orthogonal basis functions. The geodesic distance between two points along the unit sphere manifold is longer than the corresponding Hilbert distance. Accordingly, when generalizing traditional dictionary learning or sparse coding algorithms to a sphere manifold, one of the key challenges to be resolved is to exploit the manifold geometry. The reason is that in  Hilbert or Euclidean space (in the cases of finite dimension), the global linear structure can make sure the data synthesized from the atoms is contained in the same space, whereas  on the sphere manifold, the manifold geometry provides only local linear structures using the Log-Euclidean metrics \cite{ArsignyFillardPennecAyache2006}.

By taking advantage of this local linear property, a novel nonlinear dictionary learning framework \cite{XieHoVemuri2013} is proposed for data lying on the manifold of square root densities and applied to the reconstruction of DP fields given a multi-shell diffusion MRI data set \cite{SunXieYeHoEntezariBlackbankVemuri2013}. Under this paradigm, the loss of global linear structure is compensated by the local linear structure given by the tangent space using Riemannian exponential and logarithm maps. In particular,  the $l_1$-norm regularization is employed to produce sparse solutions with many zeros.  However, this sparsity scheme does not consider the global structure of data, which is very important in some computer vision tasks, especially in classification and clustering applications. Moreover, it is not  robust to noise in data.

In contrast, low rank representation \cite{LiuLinYanSunYuMa2013} uses holistic constraints as its sparse representation condition, which can reveal the latent sparse structure property embedded in a data set in high dimensional space.The lowest-rank criterion can enforce to correct corruptions and  could be robust to noise. Low rank representations on non-Euclidean geometry have received comparatively little attention. Recently, a low rank representation on Grassmann manifolds has been proposed in \cite{wanghugaosunyin2014} by mapping the Grassmann manifold onto the Euclidean space of symmetric matrices. The authors of \cite{FuGaoHongTien2015} introduced a  low rank representation for symmetric positive definite matrices measured by the extrinsic distance defined by the metric on tangent spaces and the nuclear norm simultaneously.  Some researchers  exploited the local manifold structure of the data by adding a manifold regularization term characterized by a Laplacian graph \cite{liuchezhangxu2014} or manifold matrix factorization \cite{zhangzhao2013}. However, the loss functions are still formulated in the Euclidean space. Others generalized Riemannian geometry to the matrix factorization with fixed-rank cases \cite{EldarNeedellPlan2012,shalitweinshallchechik2012,vandereyckenasilvandewalle2013}.

To our best knowledge, none of the existing work is specialized for the low rank representation on the manifold of square root densities measured by the Riemmanian distance and nuclear norm simultaneously, which motivates our study. The contribution of this work is threefold:
\begin{itemize}
  \item We propose a novel LRR model on the manifold of square root densities. The approximation quality is measured by the extrinsic distance defined by the metric on tangent spaces and the intrinsic geodesic distance on the manifold simultaneously.
  \item  We describe a simple and effective approach for optimizing our objective function as the objective function is a composition of a quadratic term, a linear term and the nuclear norm term.
  \item We present results on a few computer vision tasks on several image sets to demonstrate our new LRR model's superior performance in classification, clustering and noise robustness over other state-of-the-art methods.
\end{itemize}
\section{Preliminaries}

A manifold $\mathcal{M}$ of dimension $m$ \cite{Lee2003} is a topological space that locally resembles a Euclidean space $\mathbb{R}^m$  in a neighbourhood of each point $\mathbf x \in \mathcal{M}$. For example, lines and circles are $1D$ manifolds, and surfaces, such as a plane, a sphere, and a torus, are $2D$ manifolds. Geometrically, a tangent vector is a vector that is tangent to a manifold at a given point $\mathbf x$. Abstractly, a tangent vector is a function, defined on the set of all the smooth functions over the manifold $\mathcal{M}$, which satisfies the Leibniz differential rule. All the possible tangent vectors at $x$ constitute a Euclidean space, named the \textit{tangent space} of $\mathcal{M}$ at $x$ and denoted by $T_{\mathbf x}\mathcal{M}$. If we have a smoothly defined metric (inner-product) across all the tangent spaces $\langle\cdot,\cdot\rangle_{\mathbf x}:T_{\mathbf x}\mathcal{M}\times T_{\mathbf x}\mathcal{M} \rightarrow \mathbb{R}$  on every point $\mathbf x \in \mathcal{M}$, then we call $\mathcal{M}$ \textit{Riemannian manifold}. With a globally defined differential structure, manifold $\mathcal{M}$ becomes a differentiable manifold. The $m$-dimensional \textit{sphere} denoted by $\mathcal{S}^m$ is a specific Riemannian manifold, which has unit radius and is centered at the origin of the $m+1$ dimensional Euclidean space.  A \textit{geodesic} $\gamma :[0,1]\rightarrow \mathcal{M}$ is a smooth curve with a vanishing covariant derivative of its tangent vector field, and in particular, the Riemmannian distance     between two points $\mathbf x_i, \mathbf x_j \in \mathcal{M}$ is  the shortest smooth path connecting them on the manifold, that is the infimum of the lengths of all geodesics joining $\mathbf x_i$ and $\mathbf x_j$.

There are predominantly two operations for computations on the Riemannian manifold, namely (1) the exponential map at point $\mathbf x$, denoted by $\exp_{\mathbf x}: T_{\mathbf x}\mathcal{M} \rightarrow \mathcal{M}$,  and (2) the logarithmic map, at point $\mathbf x$, $\log_{\mathbf x}: \mathcal{M}\rightarrow T_{\mathbf x}\mathcal{M} $. The former projects a tangent vector in the tangent space onto the manifold, the latter does the reverse. Locally both mappings are diffeomorphic. Note that these maps depend on the manifold point $\mathbf x$ at which the tangent spaces are computed. Given two points $\mathbf x_i, \mathbf x_j \in \mathcal{M}$ that are close to each other,  the distance between them can be calculated through the following formula as the norm in tangent space.
\begin{equation}
\text{dist}_{\mathcal{M}}(\mathbf x_i, \mathbf x_j)=\|\log_{\mathbf x_i}(\mathbf x_j)\|_{\mathbf x_i}
\end{equation}
The squared distance function $\text{dist}^2_{\mathcal{M}}(\mathbf x,\cdot)$ is a  smooth function for all $\mathbf x \in \mathcal{M}$.

\section{Problem Formulation}
The Euclidean LRR model uses the Frobenius norm based on  metric to model the reconstruction error,  as shown in Eq. \eqref{lrrg}(if $q=$1). However, in many real world applications,   high dimension data have a Riemannian manifold structure, such as face recognition \cite{PangYuanLi2008} and object detection \cite{MaSuJurie2012}. In ideal scenarios, error should be measured according to the manifold's geometry.  Inspired by the LRR formulation for data on the Riemannian manifold of symmetric positive definite matrices in \cite{FuGaoHongTien2015}, the LRR model in Eq. \eqref{lrrg} can be changed to the following manifold form:
\begin{equation}\label{rlrr}
 \begin{aligned}
\min_{W} \sum_{i=1}^n\left\|\sum_{j=1}^n w_{ji}\log_{\mathbf x_i}(\mathbf x_j)\right\|^2_{\mathbf x_i}+\lambda\|W\|_{\star}\\
+ \nu  \sum_{i=1}^n \sum_{j\neq i} e^{\frac{d_g(\mathbf x_i;\mathbf x_j)}{\sigma}} |w_{ij}| \\
\text{s.t. } \sum_{j=1}^nw_{ji}=1, i=1,2,\ldots,n
  \end{aligned}
\end{equation}
where $\nu >0$ is a penalty parameter, $\sigma$ is a distance threshold (default:$\sigma=$1), the $i$-th column of  matrix $W$ is $\mathbf w_i=(w_{1i},\cdots, w_{ni})^T$, denoting the low rank representation for $\mathbf x_i$, and $d_g(\mathbf x_i;\mathbf x_j)$ denotes the geodesic distance between the two points on the Riemannian manifold.

\begin{figure*}[htp]
\centering
\includegraphics[width=0.9\linewidth]{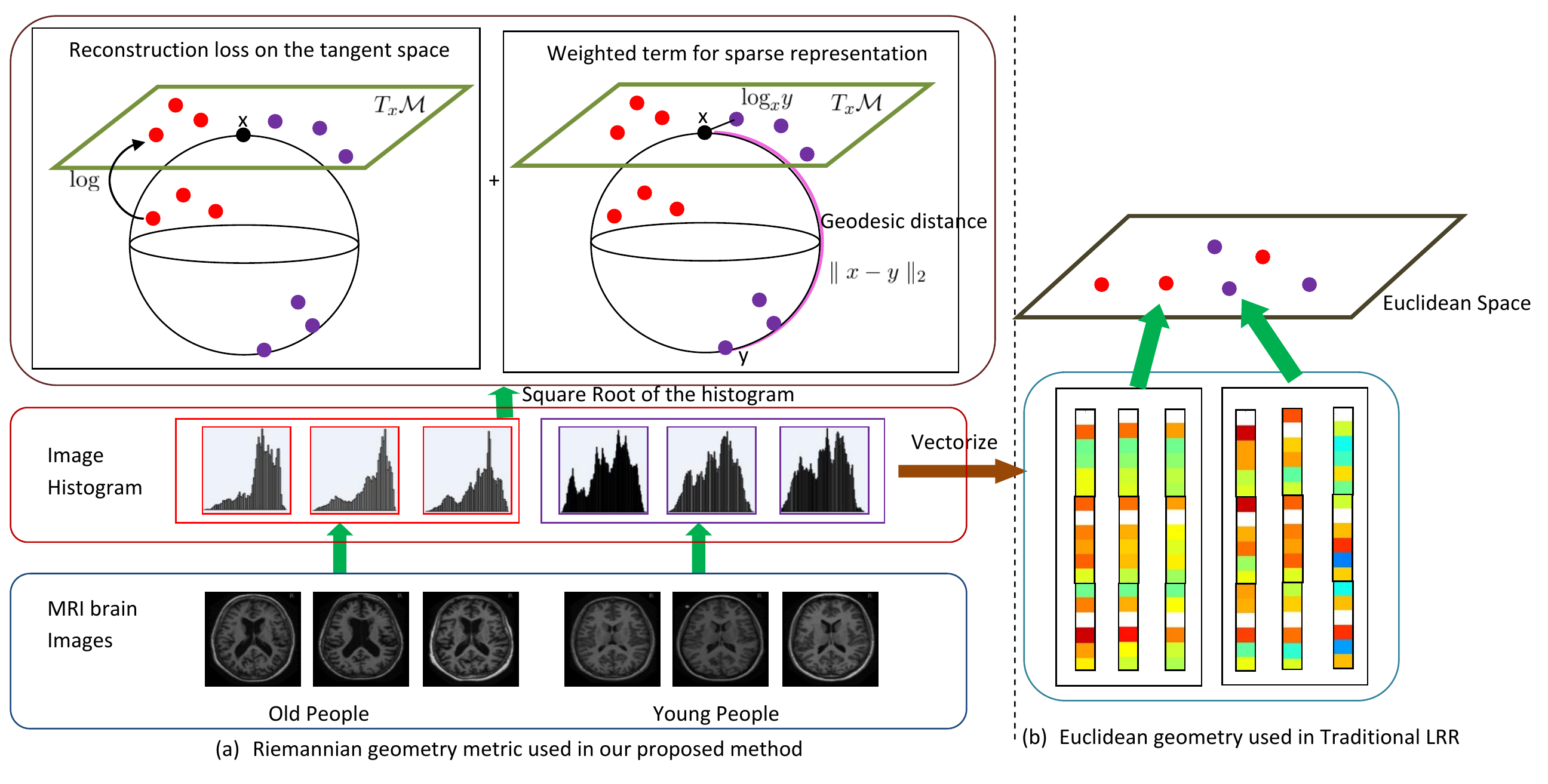}
\caption{The illustration of distance metrics used in our proposed method and Euclidean LRR methods: the purple curve denotes the manifold geodesic distance between data points $\mathbf x$ and $\mathbf y$, the black line denotes their corresponding distance on the tangent space. Without the weight term $e^{\frac{d_g(\mathbf x_i;\mathbf x_j)}{\sigma}}$, the weight between $\mathbf x$ and $\mathbf y$ would be assigned a large value. Therefore, this weight term is used to reduce the weights between far-away points.} \label{distancecom}
\end{figure*}

There is an intuitive explanation for the formulation in Eq. \eqref{rlrr}. For each point $\mathbf x_i$ on the manifold, $\mathbf x_i$ can be projected onto 0 tangent vector of tangent space at $\mathbf x_i$, other points $\mathbf x_j$ are projected as $\log_{\mathbf x_i}(\mathbf x_j)$. The norm in the first term of \eqref{rlrr} means the distance between 0 tangent vector and the linearly combined tangent vector from all the other projected tangent vectors. The weight term  $e^{\frac{d_g(\mathbf x_i;\mathbf x_j)}{\sigma}}$ for sparse representation is used to avoid assigning large weights to the far-away points in the manifold's embedding space (see further explanation in Figure.\ref{distancecom}(a)).  Therefore,  minimization aims at finding an appropriate linear combination for the ``best'' approximation in terms of tangent vectors, and meanwhile,  effectively avoiding sparse representations via far-away points that are unrelated to the local manifold structure.

The difference between the Euclidean LRR and our proposed method is illustrated in Figure \ref{distancecom}. Given a toy MR brain image dataset of 6 people from young  and old groups respectively,  the goal is to classify these sample images according to their age group.  To start with, we extract a histogram vector for each image, the square root of which lies on a finite dimensional sphere manifold. Traditional LRR simply regards the histograms as conventional feature vectors, and measures data distance using Euclidean geometry as shown in Figure~\ref{distancecom}(b). In contrast, our proposed method uses the square root of the histogram on the Riemannian manifold as input features, and the distance between input feature vectors on the manifold is approximated by the Euclidean distance between their corresponding mapped points on the tangent space, as shown in Figure~\ref{distancecom}(a). In Figure~\ref{distancecom}, we note that two points on the manifold having a large geodesic distance separating them may be near each other in the Euclidean space, which demonstrates that the important intrinsic properties of the data manifold cannot be captured using the Euclidean geometry. However, using Riemannian exponential and logarithm maps alone may fail to detect the local structure at point $\mathbf x$ in the manifold setting, because two far-away points may have a short distance on the tangent space. Accordingly, we use a term  $e^{\frac{d_g(\mathbf x_i;\mathbf x_j)}{\sigma}}$ to adjust weight values, thereby increasing the effect of nearby points in addition to their sparsity.

When the underlying Riemannian manifold is the square root densities  in the $m$-dimensional space $\mathcal{S}^m$, the geodesic distance $d_g$ between two points $\mathbf x, \mathbf y \in \mathcal{S}^m $  is simply the great circle distance between two points, which is defined formally as
\begin{equation} \label{spheregeo}
d_g(\mathbf x;\mathbf y)=\arccos(\mathbf x^T\mathbf y)
\end{equation}
where $\arccos:[-1,1]\rightarrow [0,\pi]$ is the usual inverse cosine function.

The tangent space at $\mathbf x\in \mathbb{R}^{m+1}$ is
\begin{equation}
T_\mathbf x(\mathbb{R}^{m+1})=\{\mathbf z \in \mathbb{R}^{m+1}: \mathbf z^T\mathbf x=0 \}
\end{equation}

Then the geodesic started at $\mathbf x$ in the direction $\mathbf v\in T_\mathbf x(\mathbb{S}^m)$ is given by the formula
\begin{equation}
\gamma(t)=\cos(t)\mathbf x+\sin(t)\frac{\mathbf v}{\mid \mathbf v \mid}
\end{equation}
and the exponential and logarithm maps are given by
\begin{equation}
\exp_\mathbf x(\mathbf v)=\cos(|\mathbf v|)\mathbf x+\sin(|\mathbf v|)\frac{\mathbf v}{ \mid \mathbf v \mid}
\end{equation}
\begin{equation}
\log_\mathbf x(\mathbf y)=\mathbf u\arccos(\mathbf x^T\mathbf y)/\sqrt{\mathbf u^T\mathbf u}
\end{equation}
where $\mathbf u=\mathbf y-(\mathbf x^T\mathbf y)\mathbf x$.

Hence the LRR on the manifold of unit sphere is defined by
\begin{equation} \label{spherelrr1}
\begin{aligned}
\min_W \frac{1}{2}\sum_{i=1}^n\sum_{j=1}^n \left|\left|\d w_{ji}\cos^{-1}(\langle \mathbf x_i,\mathbf x_j \rangle)\frac{\mathbf u_{ij}}{|\mathbf u_{ij}|}\right|\right|_{\mathbf x_i}^2+\lambda ||W||_{\star}\\
+ \nu  \sum_{i=1}^n \sum_{j\neq i} e^{\frac{d_g(\mathbf x_i;\mathbf x_j)}{\sigma}} |w_{ij}| \\
\text{s.t. } \sum_{j=1}^nw_{ji}=1, i=1,2,\ldots,n\\
\end{aligned}
\end{equation}
where $\mathbf u_{ij}=\mathbf x_j-\langle \mathbf x_i, \mathbf x_j \rangle \mathbf x_i$. The metric here is the inner product inherited from the standard inner product on $\mathbb{R}^m$ given by
\begin{equation}
\langle \xi, \eta\rangle_\mathbf x=\xi^T\eta
\end{equation}
Accordingly, the LRR problem can be re-written as
\begin{equation} \label{spherelrr2}
\begin{aligned}
\min_W \frac{1}{2}\sum_{i=1}^n\sum_{j=1,k=1}^n  w_{ji}w_{ki}\cos^{-1}(\langle \mathbf x_i,\mathbf x_j \rangle)\cos^{-1}(\langle \mathbf x_i,\mathbf x_k \rangle)\frac{\mathbf u_{ij}^T\mathbf u_{ik}}{|\mathbf u_{ij}|\cdot|\mathbf u_{ik}|}\\
+\lambda ||W||_{\star} + \nu  \sum_{i=1}^n \sum_{j\neq i} e^{\frac{d_g(\mathbf x_i;\mathbf x_j)}{\sigma}} |w_{ij}| \\
\text{s.t. } \sum_{j=1}^nw_{ji}=1, i=1,2,\ldots,n\\
\end{aligned}
\end{equation}
where $|\mathbf u_{ij}|=\sqrt{1-\langle \mathbf x_i,\mathbf x_j\rangle^2}$ and $\mathbf u_{ij}^T\mathbf u_{ik}=\langle \mathbf x_i,\mathbf x_k\rangle-\langle \mathbf x_i,\mathbf x_j\rangle \langle \mathbf x_i,\mathbf x_k\rangle$. Problem \eqref{spherelrr2} can be reformulated into the following problem
\begin{equation}\label{spherelrr3}
\begin{aligned}
\min_W \frac{1}{2}\sum_{i=1}^n \mathbf w_{i}^TQ_i\mathbf w_i+\lambda \|W\|_{\star}+\nu  \sum_{i=1}^n \sum_{j\neq i} e^{\frac{d_g(\mathbf x_i;\mathbf x_j)}{\sigma}} |w_{ij}|\\
\text{s.t. }  \sum_{j=1}^nw_{ji}=1, i=1,2,\ldots,n
\end{aligned}
\end{equation}
where
\[
Q_i=\left[\cos^{-1}(\langle \mathbf x_i,\mathbf x_j\rangle)\cos^{-1}(\langle \mathbf x_i,\mathbf x_k\rangle)\frac{\mathbf u_{ij}^T\mathbf u_{ik}}{|\mathbf u_{ij}|\cdot|\mathbf u_{ik}|}\right]
\] is a $n\times n$ matrix, and the $(j,k)$th element of $Q_i$ is\\
$\cos^{-1}(\langle \mathbf x_i,\mathbf x_j\rangle)\cos^{-1}(\langle \mathbf x_i,\mathbf x_k\rangle)\frac{\mathbf u_{ij}^T\mathbf u_{ik}}{|\mathbf u_{ij}|\cdot|\mathbf u_{ik}|}$.

\section{Solution to LRR on the manifold of Square Root Densities}
In this section, we consider an algorithm to solve the constrained optimization problem in Eq. \eqref{spherelrr3}. We propose to use the Augmented Lagrange Multiplier (ALM) method \cite{ShenWenZhang2014} to solve it.  The reason we choose the ALM to solve this optimization problem is threefold:
(1) superior convergence property of ALM makes it very attractive; (2) parameter tuning is  much easier than the iterative thresholding algorithm \cite{LinChenMa2010}; and (3) it converges to an exact optimal solution.

First of all, the augmented Lagrange problem of (\ref{spherelrr3}) can be written as
\begin{equation}\label{alp}
\begin{aligned}
L=&\sum_{i=1}^n \left(\frac{1}{2}\mathbf w_{i}^TQ_i\mathbf w_i+\nu  \sum_{j\neq i} e^{\frac{d_g(\mathbf x_i;\mathbf x_j)}{\sigma}}|w_{ji}|+y_i(\sum_{j=1}^nw_{ji}-1)\right.\\
&+\left.\frac{\beta}{2}(\sum_{j=1}^nw_{ji}-1)^2\right)+\lambda ||W||_{\star}
\end{aligned}
\end{equation}
where $y_i$ are Lagrangian multipliers, and $\beta$ is a weight to tune the error term of $(\sum_{j=1}^nw_{ji}-1)^2$.

In fact, the above problem can be solved by updating one variable at a time with all the other variables fixed. More specifically,
the iterations of ALM go as follows:

1) Fix all others to update $W$: We define a function $F(W)$ by
\begin{equation}\label{fw}
\begin{aligned}
F(W)=&\sum_{i=1}^n  \left(\frac{1}{2} \mathbf w_{i}^TQ_i\mathbf w_i+\nu  \sum_{j\neq i} e^{\frac{d_g(\mathbf x_i;\mathbf x_j)}{\sigma}}|w_{ji}|+y_i(\sum_{j=1}^nw_{ji}-1)\right.\\
&+\left.\frac{\beta}{2}(\sum_{j=1}^nw_{ji}-1)^2\right)
\end{aligned}
\end{equation}
and it is easy to prove that
\begin{equation}
\frac{\partial F}{\partial \mathbf w_i}=Q_i\mathbf w_i+\nu  \sum_{j\neq i} e^{\frac{d_g(\mathbf x_i;\mathbf x_j)}{\sigma}}+y_i\textbf{1}+\beta(\sum_{j=1}^nw_{ji}-1)\textbf{1}
\end{equation}
where $\textbf{1}\in \mathbb{R}^n$ is a column vector of all ones. Now at the current location $W^{(k)}$, we take a linearization of $F(W)$,
\begin{equation}
\begin{aligned}
F(W)\approx & F(W^{(k)})+\langle\partial F(W^{(k)}), W-W^{(k)}\rangle \\
& +\frac{\mu_k}{2}||W-W^{(k)}||_F^2
\end{aligned}
\end{equation}
where $\partial F(W^{(k)})$ is a gradient matrix whose $i$-th row is given by
\begin{equation}\label{partialfw}
Q_i\mathbf w_i^{(k)}+\nu  \sum_{j\neq i} e^{\frac{d_g(\mathbf x_i;\mathbf x_j)}{\sigma}}+y_i^{(k)}\textbf{1}+\beta(\sum_{j=1}^nw_{ji}^{(k)}-1)\textbf{1}
\end{equation}
Taking Eq.\eqref{partialfw} into Eq.\eqref{alp}, we have
\begin{align}
W^{(k+1)}=&\arg\min_{W} F(W^{(k)})+\langle\partial F(W^{(k)}), W-W^{(k)}\rangle \notag\\
& +\frac{\mu_k}{2}||W-W^{(k)}||_F^2+\lambda ||W||_{\star} \label{almw}\\
=& \arg\min_{W}\frac{\mu_k}{2}||W-\big(W^{(k)}-\frac{1}{\mu_k}\partial F(W^{(k)})\big)||\notag\\
&+\lambda ||W||_{\star} \notag
\end{align}
The above problem admits a closed form solution by using SVD thresholding operator to $A=W^{(k)}-\frac{1}{\mu_k}\partial F(W^{(k)})$. Taking SVD for $A=U\Sigma V^T$, then the new iteration is given by
\begin{equation}
W^{(k+1)}=U \text{soft}(\Sigma, \frac{\mu_k}{\lambda})V^T
\end{equation}
where $\text{soft}(\Sigma,\sigma)=\max\{0,(\Sigma_{ii}-\frac{1}{\sigma})\}$ is the soft thresholding operator for a diagonal matrix, see \cite{CaiCandasShen2010}.

2) Fix all others to update $y_i$ by
\begin{equation}
y^{k+1}_i\leftarrow y^k_i+\beta_k(\sum_{j=1}^n w^{k+1}_{ji}-1) \label{UpdateY}
\end{equation}
After a number of iterations by alternately updating $W$ and $y_i$ respectively,  we achieved a complete solution to LRR on the manifold of square root densities. The whole procedure of LRR on the manifold of square root densities is summarized in Algorithm \ref{agalm}.
\begin{algorithm}[ht]
\caption{Solving Problem (\ref{spherelrr3}) by ALM}
\label{agalm}
\begin{algorithmic}[1]
\REQUIRE \text{The square root densities sample set}$\{\mathbf x_i\}_{i=1}^n, \mathbf x_i \in \mathcal{S}^m$, \text{penalty parameters} $\lambda$ \text{and} $\nu$, \text{and a distance threshold} $\sigma=$1, the initial $W$.
\ENSURE: The low rank representation $W$
\STATE initialize: $y_i=0$ ($i=1,\ldots,n), \beta=10^{-6}, \beta_{\max}=10^{10}, \rho=1.1, \varepsilon=10^{-8}$.
\STATE $\mathbf u_{ij}\leftarrow \mathbf x_j-\langle \mathbf x_i, \mathbf x_j \rangle \mathbf x_i$ for $i,j=1,2,\ldots, n$;
\STATE $Q_i\leftarrow \left[\cos^{-1}(\langle \mathbf x_i,\mathbf x_j\rangle)\cos^{-1}(\langle \mathbf x_i,\mathbf x_k\rangle)\frac{\mathbf u_{ij}^T\mathbf u_{ik}}{|\mathbf u_{ij}|\cdot|\mathbf u_{ik}|}\right]$ for $i,j,k=1$ to $n$;
\STATE $d_g(\mathbf x_i;\mathbf x_j)\leftarrow$ \text{Eq.}\eqref{spheregeo};
\WHILE{$|\sum_{j=1}^nw_{ji}-1| \geq \varepsilon$ }
     \STATE $W$ is computed by solving the problem \eqref{almw} using SVD thresholding operator;
     \STATE $y_i$ is updated by \eqref{UpdateY}.
     \STATE $\beta\leftarrow \min(\rho\beta,\beta_{max})$
\ENDWHILE
\end{algorithmic}
\end{algorithm}

\subsection{Convergence Analysis}
The convergence of ALM algorithm has been guaranteed by  Theorem 3 in \cite{LinLiuSu2011}.  To ensure the convergence of the Algorithm \ref{agalm}, we only need prove that problem (\ref{spherelrr3}) is a convex optimization problem.

\textbf{Theorem 1} Each $Q_i$ is a semi-positive definite matrix, thus problem (\ref{spherelrr3}) is a convex optimization.

\textbf{Proof}. The $jk$-th element $q_{jk}^i$ of $Q_i$ can be represented by
\begin{equation}
\begin{aligned}
q_{jk}^i&=\cos^{-1}(\langle \mathbf x_i,\mathbf x_j\rangle)\cos^{-1}(\langle \mathbf x_i,\mathbf x_k\rangle)\frac{\mathbf u_{ij}^T\mathbf u_{ik}}{|\mathbf u_{ij}|\cdot|\mathbf u_{ik}|}\\
&=\cos^{-1}(\langle \mathbf x_i,\mathbf x_j\rangle)\frac{\mathbf u_{ij}^T}{|\mathbf u_{ij}|}\cos^{-1}(\langle \mathbf x_i,\mathbf x_k\rangle)\frac{\mathbf u_{ik}}{|\mathbf u_{ik}|}\\
&=\mathbf l_{ij}^T \mathbf l_{ik}\\
\end{aligned}
\end{equation}
where $\mathbf l_{ij}=\cos^{-1}(\langle \mathbf x_i,\mathbf x_j\rangle)\frac{\mathbf u_{ij}}{|\mathbf u_{ij}|}$.

If we define a new matrix $V_i$ by
\begin{equation}
V_i=[\mathbf l_{i1},\mathbf l_{i2}, \ldots,\mathbf l_{in}]
\end{equation}
then we can see $Q_i=V_i^TV_i$, which means $Q_i$ has a decomposition as the product of a matrix with its transpose, hence $Q_i$ is semi-positive definite.

\subsection{Algorithm Complexity}
For ease of analysis, we assume that  $\mathbf x_i$ is a $m$-dimensional square root densities sample, and the iteration number of ALM algorithm is  $s$. The complexity of Algorithm \ref{agalm} can be decomposed into two parts: the data preparation part (Steps 1-4) and the ALM solution part (Steps 5-9).

In the data preparation part, $\mathbf u_i$ and $\parallel \mathbf x_i-\mathbf x_j\parallel_{2}$ for $(i,j=1,\ldots, n)$ can be computed with a complexity of $O(n^2m^2)$. The bottleneck of the data preparation procedure is the computation of all $Q_i$. As we could store the inner product values of $\mathbf x_i$ and $\mathbf x_j$ computed in Step 2, the time complexity of computing $Q_i$ should be $O(n^3)$. As $Q_i = V^T_iV_i$, to break this complexity bound, we use a parallel matrix multiplication scheme \cite{ChenWangLu1992} to reduce the complexity to $O(n\text{log}n)$. Therefore, the  complexity of data preparation is $O(n^2m^2)+O(n\text{log}n)$.

In the ALM, the major computation cost is for SVD of an $n\times n$ matrix in Step 6 with a complexity of $O(n^3)$. Fortunately, we can utilize the accelerated method in \cite{LinLiuSu2011} to reduce the complexity of partial SVD computation to $O(rn^2)$, where $r$ is predicted rank of $W^{(k+1)}$. For $s$ iterations, the complexity of ALM solution is $O(srn^2)$.

With the above analysis, the overall complexity of Algorithm \ref{agalm} is  given as
\begin{equation}
O(n^2m^2)+O(n\text{log}n)+O(srn^2)
\end{equation}
 As $m \ll n$, the complexity of Algorithm \ref{agalm} mainly depends on the size of data set $n$. It can be approximated as $O(n^2m^2)+O(srn^2)$, which is similar to the complexity of the original LRR algorithm $O(srn^2)$.
\section{Experiments} To evaluate the proposed LRR model on the manifold of square root densities, we apply it to both clean and corrupted image datasets for image classification and image segmentation.

To apply our method for image classification, we firstly employ our proposed LRR model to obtain the low rank features, and the classification accuracy is computed by training a SVM classifier on the low rank features. In order to compare performances with the state-of-the-art methods, we use the following four baseline methods:
\begin{itemize}
  \item SVM on vectorized data: we directly vectorize the manifold data to form their Euclidean features and train an SVM using these Euclidean features.
  \item LRR +SVM: we firstly apply the Euclidean LRR model in \cite{LiuLiuYu2010} on the Euclidean features (vectorized the manifold data) to obtain the  low rank features, and then train an SVM classifier on the low rank features.
  \item GKNN: it is a K-nearest neighbour classifier that uses the geodesic distance on the manifold for determining neighbours, and solves the classification problem on the manifold directly without LRR transforms.
  \item SC+SVM: it is a variant of our proposed method within the same framework. The only difference between SC +SVM and our proposed model is that $l_1$ regularization on $W$ is used in the objective function Eq.\eqref{spherelrr3}.
\end{itemize}
 All SVMs used in the experiments are trained using the LIBSVM package \cite{ChangLin2001}.

 To apply our method for subspace clustering, we compute the low rank features using the proposed method, and then use these low rank features as the input of the Ncut clustering algorithm.  The baselines used in the experiments are listed as follows:
 \begin{itemize}
   \item Ncut \cite{ShiMalik2000}: we simply conduct the Ncut clustering algorithm over the vectorized manifold data.
   \item LRR+Ncut: we use the Euclidean LRR method on the Euclidean features to obtain the  low rank features, and then apply Ncut to the low rank features.
   \item Geodesic Distance based Ncut (GNcut): it is a variant of the traditional Ncut method,  based on the Riemannian geodesic distance introduced in this paper as the node similarity measure.
   \item SC+Ncut: it is similar to our proposed method, except $l_1$ regularization on $W$  used in the objective function Eq.\eqref{spherelrr3}.
 \end{itemize}
\subsection{Performance for Image Classification} 
\begin{figure*}
\centering
\includegraphics[width=\linewidth]{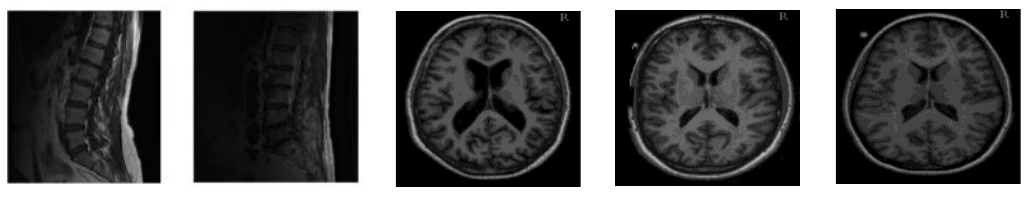}
\caption{ Sample Images for classification: From left to right, there are two sample images from the LUMBAR dataset belonging to the degenerative lumbar spine  and standard lumbar spine groups, respectively, and three sample images from the OASIS dataset belonging to the Young, Middle-aged and Old groups, respectively.} \label{figClassifydata}
\end{figure*}

In this experiment, we evaluate the performance of our proposed method in terms of image classification using OASIS \cite{MarcusWangParkerCsernanskyMorrisBuckner2007} and LUMAR \cite{DICOM} datasets.  Sample images from both data sets are shown in Figure \ref{figClassifydata}.

The OASIS dataset consists of T1-weighted MR brain images from a cross-sectional collection of 416 subjects aged 18 to 96. The subjects are all right-handed and include both men and women.  Each MRI scan has a resolution of 176$\times$208 pixels. The whole OASIS images are categorized into three groups: young subjects (younger than 40), middle-aged subjects (between 40 and 60) and old subjects (older than 60). We aim to classify each MRI image into its corresponding group. We note that the subtle differences in anatomical structure across different age groups in Figure. \ref{figClassifydata} are apparent.

The LUMBAR dataset collects  MRI scans of human lumbar spine from 60 subjects, who are randomly sampled from normal people and patients with lumbar degenerative disease groups respectively. Each image has a resolution of 643$\times$574 pixels. The classification problem is to recognize the abnormal lumbar spines from the normal ones. Two sample images from each group in the LUMBAR dataset are shown in Figure. \ref{figClassifydata}.

To generate the square root densities data, we use some pre-processing techniques  on the above two MRI datasets. First of all, we obtain a displacement field for each MRI image in the  dataset using a nonrigid group-wise registration method described in \cite{MyronenkoSong2010}. Then we compute the histogram of the displacement vector for each image as the feature for classification. In our experiment, the number of bins in each direction is set to 4, and the resulting 16-dimensional histogram is generated as the feature vector for the SVM and LRR+SVM methods. The square root of the histogram is used in the GKNN, SC+SVM and our new method.

The classification results on the above two datasets are reported in Table \ref{tb-ic}. It shows that  LRR+SVM outperforms SVM, which indicates that low rank features are more discriminative for classification than the original high-dimensional features. As GKNN considers the intrinsic property of the data manifold, its performance is superior to the counterparts using extrinsic Euclidean metric (SVM and LRR+SVM), which confirms the importance of intrinsic geometry. As we expected, our method and SC+SVM utilizing both intrinsic geometry and sparse feature transform (i.e. sparse coding and low rank representation) outperform all the other baselines. The performance of our proposed method also outperforms its variant method SC+SVM under the same framework, which demonstrates that the lowest-rank criterion is accurate for modeling the embedding manifold structures of high dimensional data.
\begin{table}[h]
\centering
\caption {Classification Accuracy comparisons on Clean data sets. For the OASIS dataset, there are three binary classifications (YM: Young vs Middle-aged, MO: Middle-aged vs Old and YO: Young vs Old), and one three-class classification (YMO: Young, Middle-aged and Old). For the LUMBAR dataset, there is one binary classification (NA:Normal vs Abnormal).}
\begin{tabular}{|c|c|c|c|c|c|}
\hline
\multicolumn{1}{|c|}{Datasets} &
\multicolumn{4}{c|}{OASIS} &
\multicolumn{1}{c|}{LUMBAR}
 \\  \hline
\multicolumn{1}{|c|}{Classes} & {YM}& {MO} & {YO} & {YMO} & {NA}\\ \hline
\multicolumn{1}{|c|}{SVM} & {89.05}& {92.58} & {93.22} & {91.36} & {87.32}\\ \hline
\multicolumn{1}{|c|}{LRR+SVM} & {90.18}& {93.24} & {94.57} & {92.58} & {89.79}\\ \hline
\multicolumn{1}{|c|}{GKNN} & {91.75}& {95.59} & {96.88} & {94.49} & {92.95}\\ \hline
\multicolumn{1}{|c|}{SC+SVM} & {94.56}& {96.77} & {97.25} & {96.17} & {95.11}\\ \hline
\multicolumn{1}{|c|}{Proposed} & {97.25}& {99.88} & {99.97} & {99.87} & {98.8}\\ \hline
\end{tabular}\label{tb-ic}
\end{table}
\subsection{Performance on Subspace Clustering}In this section, our proposed method is used to solve the subspace clustering problem. We evaluate our model on 4 hyperspectral images, including pavia centre scene \cite{FauvelBenediktssonChanussotSveinsson2008}, salinas-A scene \cite{PlazaMartinezPerezPlaza2004}, samson \cite{ZhuWangXiangFanPan2014} and jasper ridge \cite{ZhuWangXiangFanPan2014}.

The Pavia centre scene was acquired by the ROSIS sensor during a flight campaign over Pavia, nothern Italy. It is a 1096$\times$1096 pixels image, with 102 spectral bands. The geometric resolution is 1.3 meters, and the image groundtruths differentiate 9 classes.

The Salinas-A scene is a small subscene of salinas image, which was collected by the 224-band AVIRIS sensor with a spatial resolution of 3.7-meter pixels. The salinas-A scene consists of  86$\times$83 pixels located within the salinas scene at [rows, columns] = [591-676, 158-240] and includes six classes.

The Samson consists of 952$\times$952 pixels, each of which has 156 channels covering the wavelengths from 401 nm to 889 nm. The spectral resolution is highly up to 3.13 nm. To reduce the computational cost, a sub region of 95$\times$95 pixels is used, starting from the (252,332)-th pixel in the original image. There are three targets in this image, i.e., soil, tree and  water respectively.

The Jasper ridge image has 512 x 614 pixels, with 224 channels ranging from 380 nm to 2500 nm. The spectral resolution is up to 9.46nm. Since this hyperspectral image is too complex to get the ground truth, we consider a subimage of 100 $\times$ 100  pixels,  starting from the (105,269)-th pixel in the original image. In our experiments,  the channels 1-3, 108-112, 154-166 and 220-224 are removed due to dense water vapor and atmospheric effect. There are four classes in this data: road, soil, water and tree.

In order to extract the square root densities information embedded in the original hyperspectral images, we compute the histogram of the band values for each pixel. In our experiment, the number of bins is set to 6, so  the resulting 6-dimensional histogram vector is used as the pixel feature vector for the Ncut and LRR$+$Ncut methods, while the square root of the histogram vector is used in GNcut, SC+Ncut and our new method.

The subspace clustering results are detailed in Table \ref{tbcluster}. Obviously, the performance of Ncut is inferior to that of LRR+Ncut, as the low rank features are more discriminative and useful  than the data themselves for clustering problems. Similarly, the clustering accuracy of GNut is 3\% lower than that of our proposed method on average. This is mainly because GNcut directly works on the original high-dimensional manifold data, without further exploring the latent subspace structure hidden in the manifold data.  As GNcut, SC+Ncut and our proposed model use the Riemannian distance measurement, they perform better than  Ncut and LRR+Ncut, in which the Euclidean distance is used to measure the similarities between points on the manifold. We can assert that taking inherent manifold structure into account can effectively improve clustering accuracy. We also note that our proposed method has the highest accuracy among all the baseline methods, which suggests that the combination of Riemannian distance and LRR model brings good accuracy for Ncut clustering. To visualize our proposed model's effectiveness in subspace clustering, we illustrate the clustering results of the above 4 images in Figure \ref{cluster123}-\ref{cluster4}.
\begin{table}\centering
\caption{Subspace clustering accuracy on hyperspectral images}
\centering
\begin{tabular}{|l*{4}{|c}|}
\hline
 Dataset &Paviacenter &SalinasA &Samson&Jasperridge \\
 \hline
 Classes &9&6&3&4\\
\hline
Ncut& 76.85& 78.59 & 84.52 & 83.75\\
\hline
LRR+Ncut & 79.21  & 81.23  & 87.76 & 85.02\\\hline
GNcut & 83.56 & 85.97 & 90.89 & 88.46\\\hline
SC+Ncut & 85.01 &86.33 & 93.61 & 90.05\\\hline
Proposed & 88.67 & 88.99 & 97.01 & 94.71\\
\hline
\end{tabular}\label{tbcluster}
\end{table}
\begin{figure*}
\centering
\subfigure[Original Pavia Center]{
\includegraphics[width=0.3\linewidth, height=0.2\linewidth]{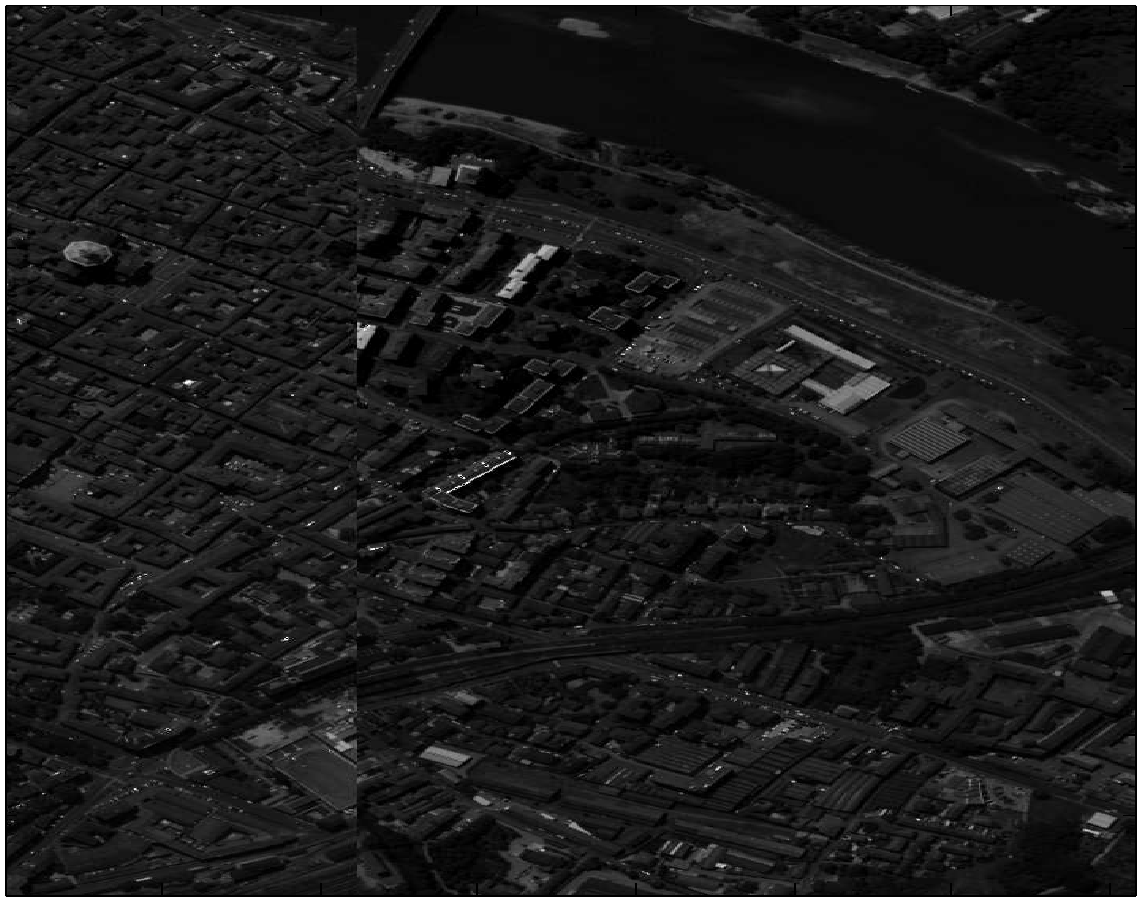}
}
 \subfigure[Ncut]{
\includegraphics[width=0.3\linewidth, height=0.2\linewidth]{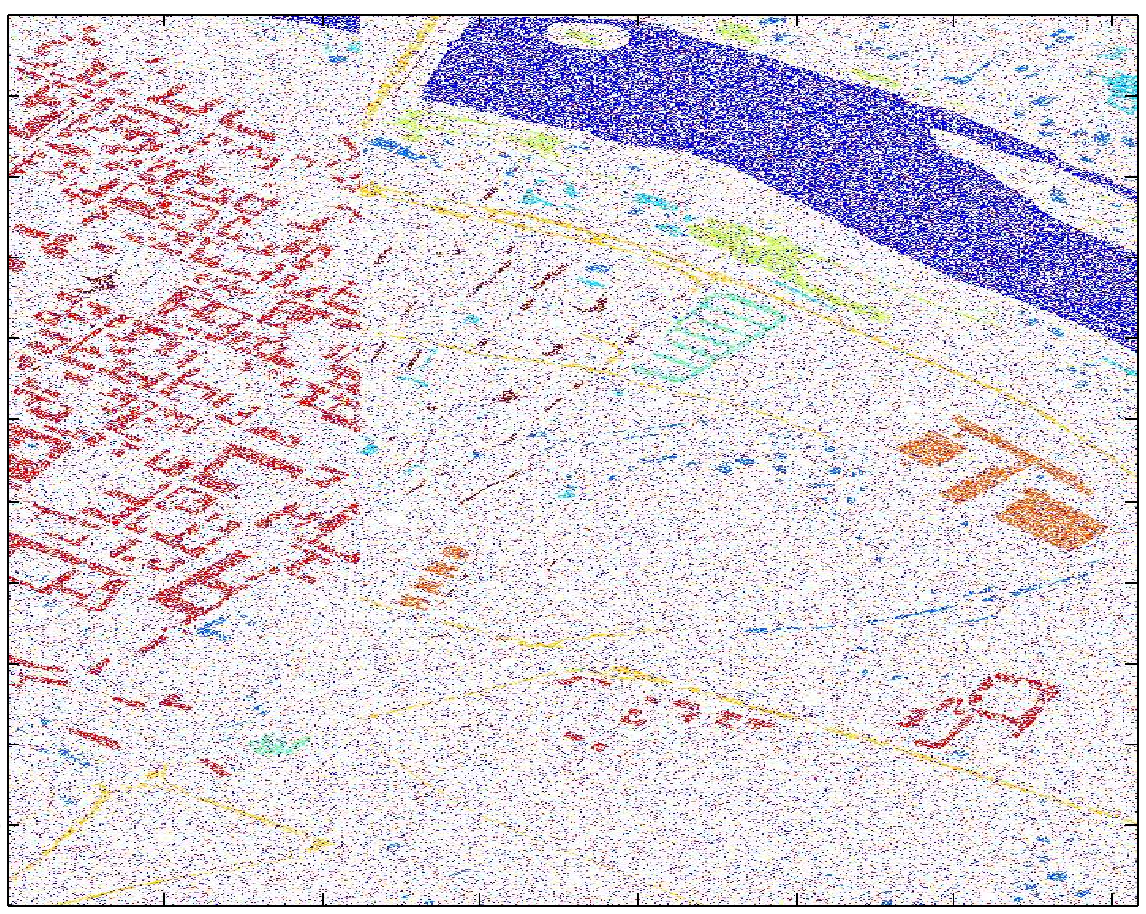}
}
 \subfigure[LRR+Ncut ]{
\includegraphics[width=0.3\linewidth, height=0.2\linewidth]{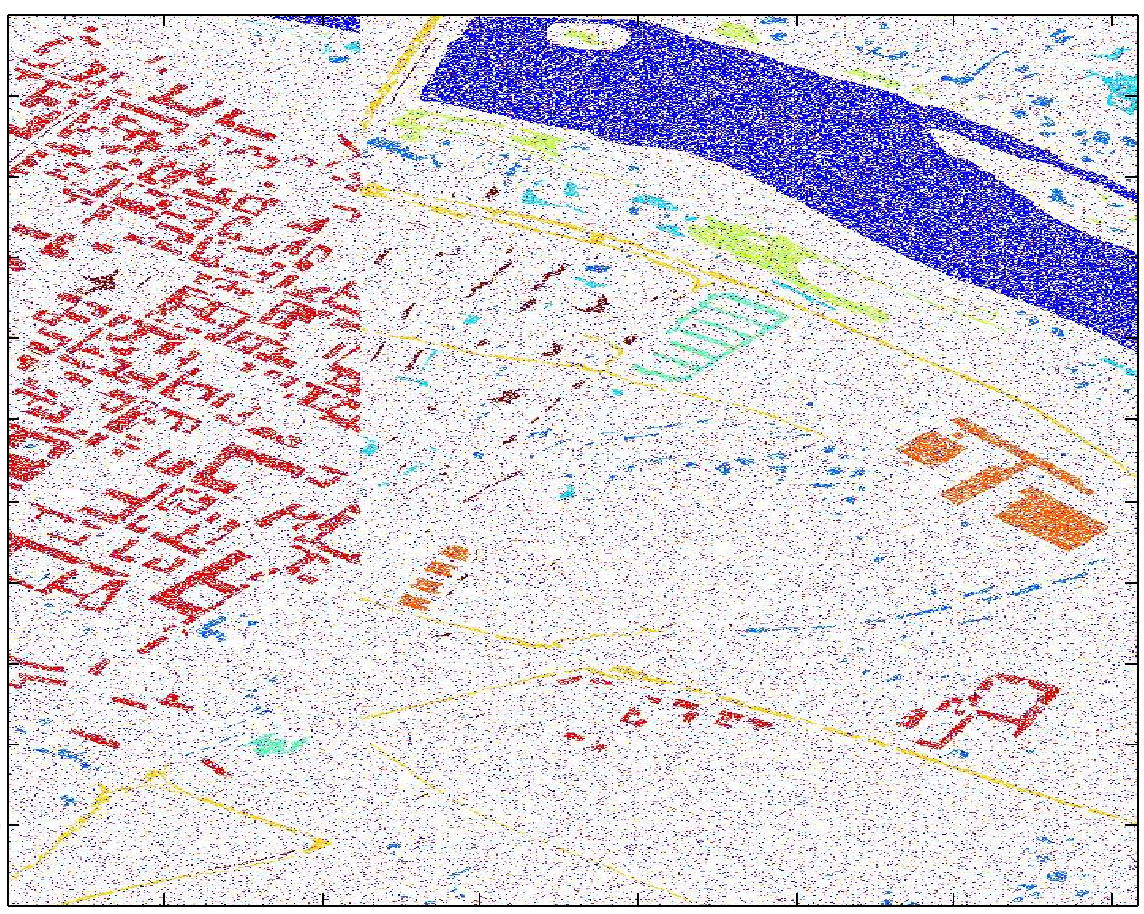}
}
 \subfigure[GNcut]{
\includegraphics[width=0.3\linewidth, height=0.2\linewidth]{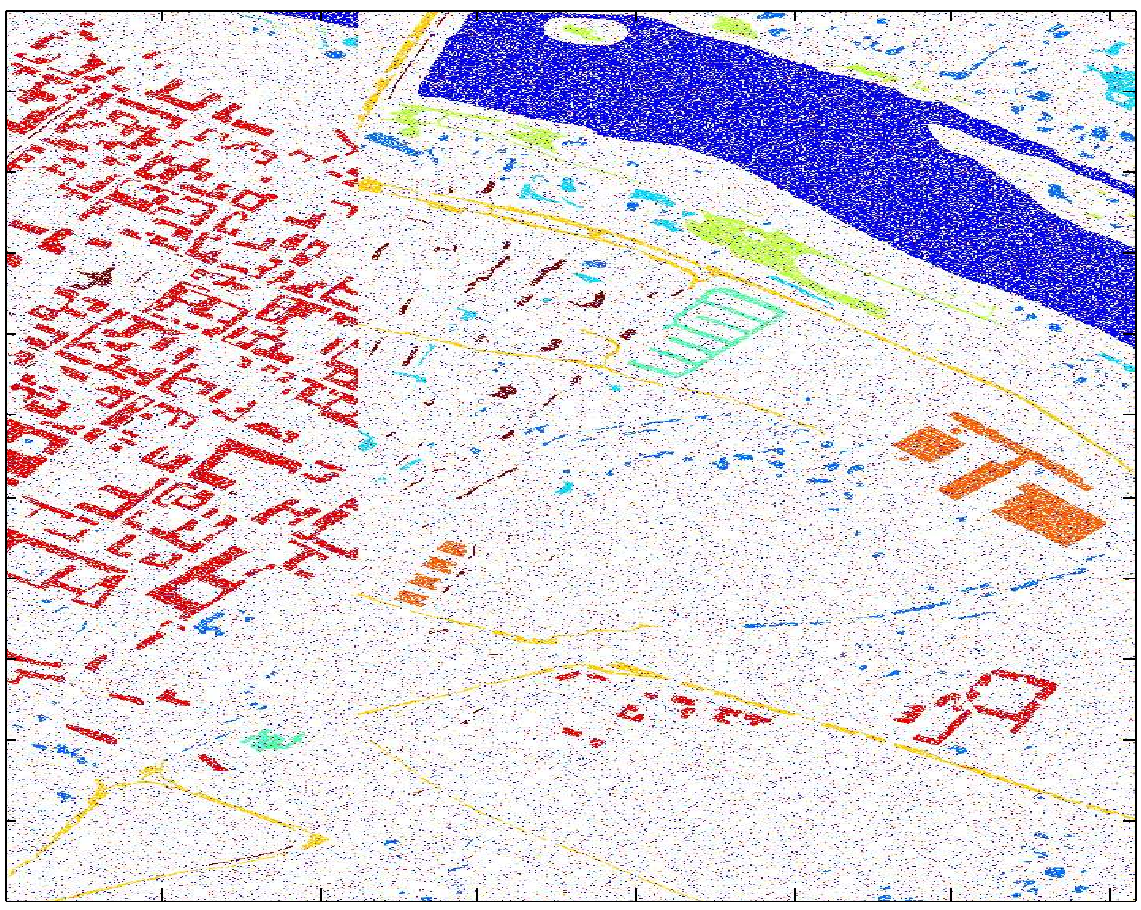}
}
 \subfigure[SC+Ncut]{
\includegraphics[width=0.3\linewidth, height=0.2\linewidth]{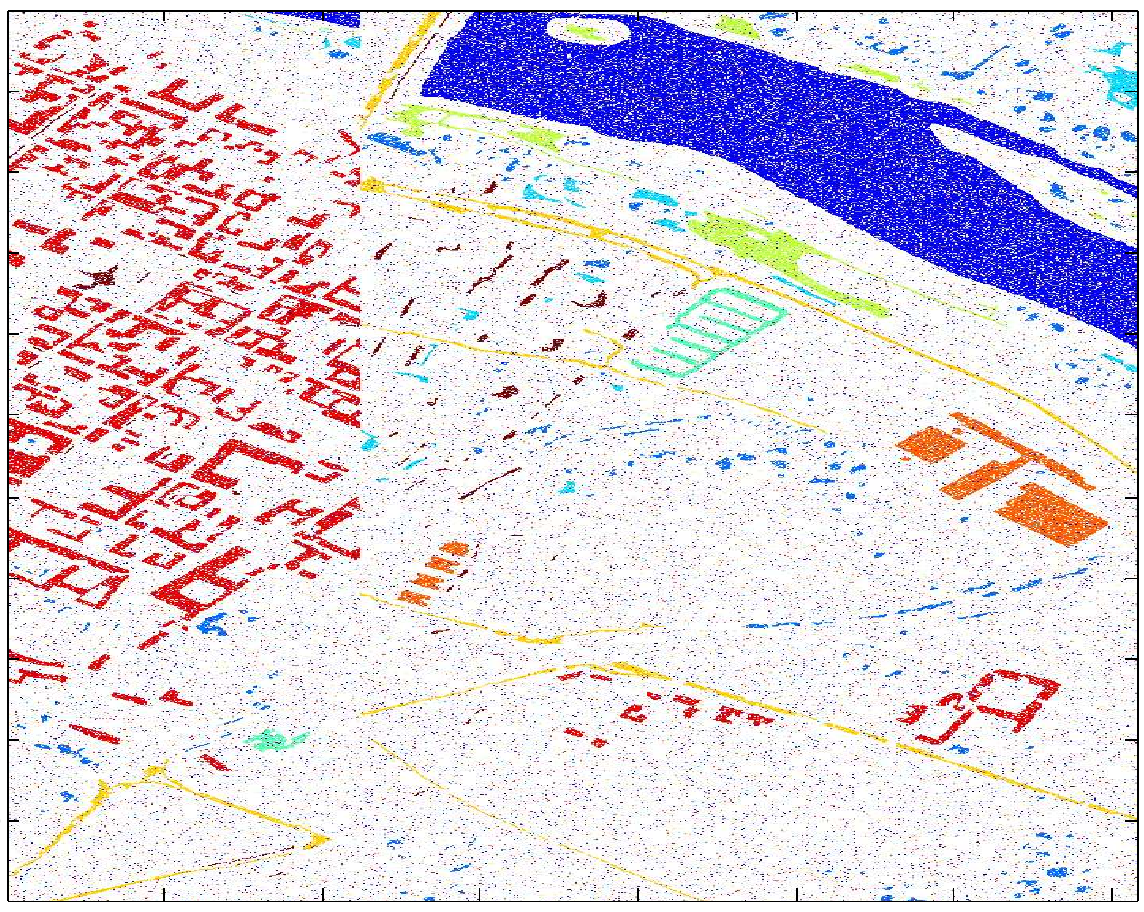}
}
 \subfigure[Proposed]{
\includegraphics[width=0.3\linewidth, height=0.2\linewidth]{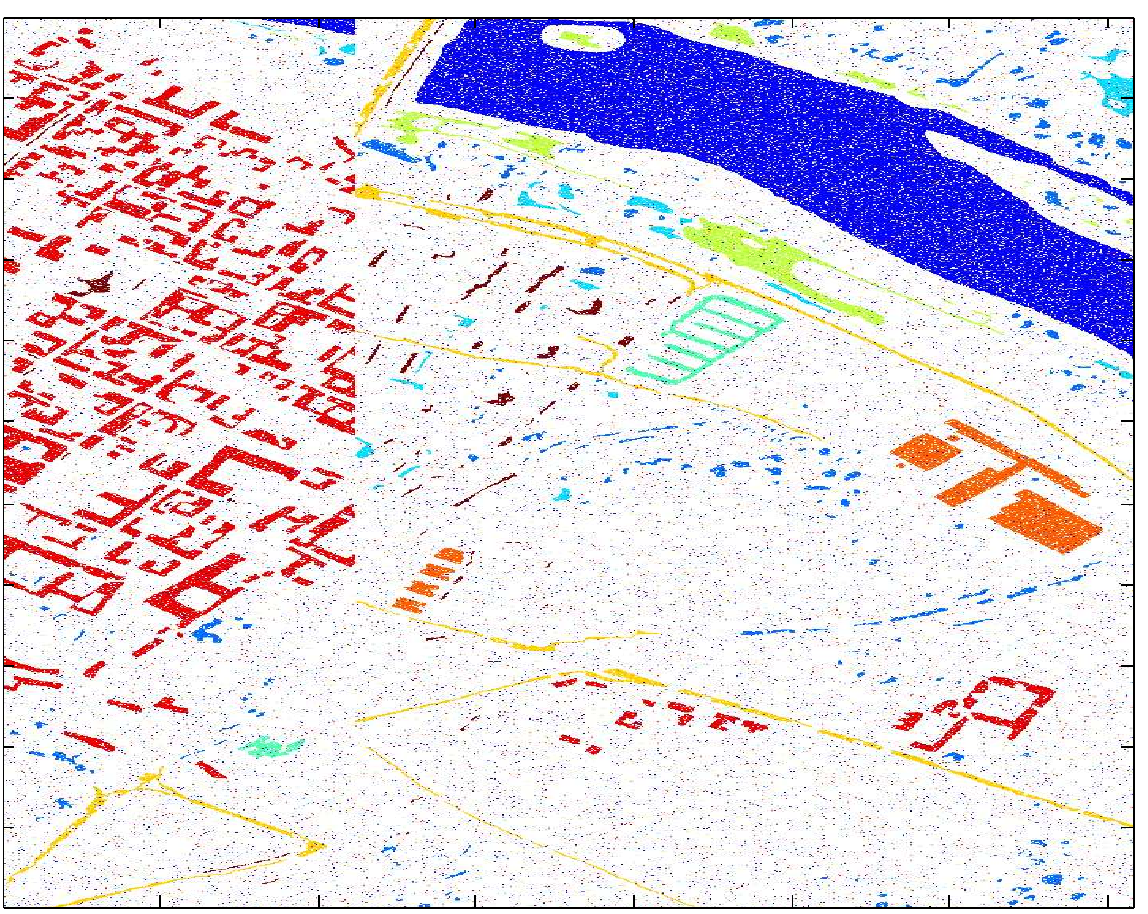}
}
 \subfigure[Original Salinas-A]{
\includegraphics[width=0.3\linewidth, height=0.2\linewidth]{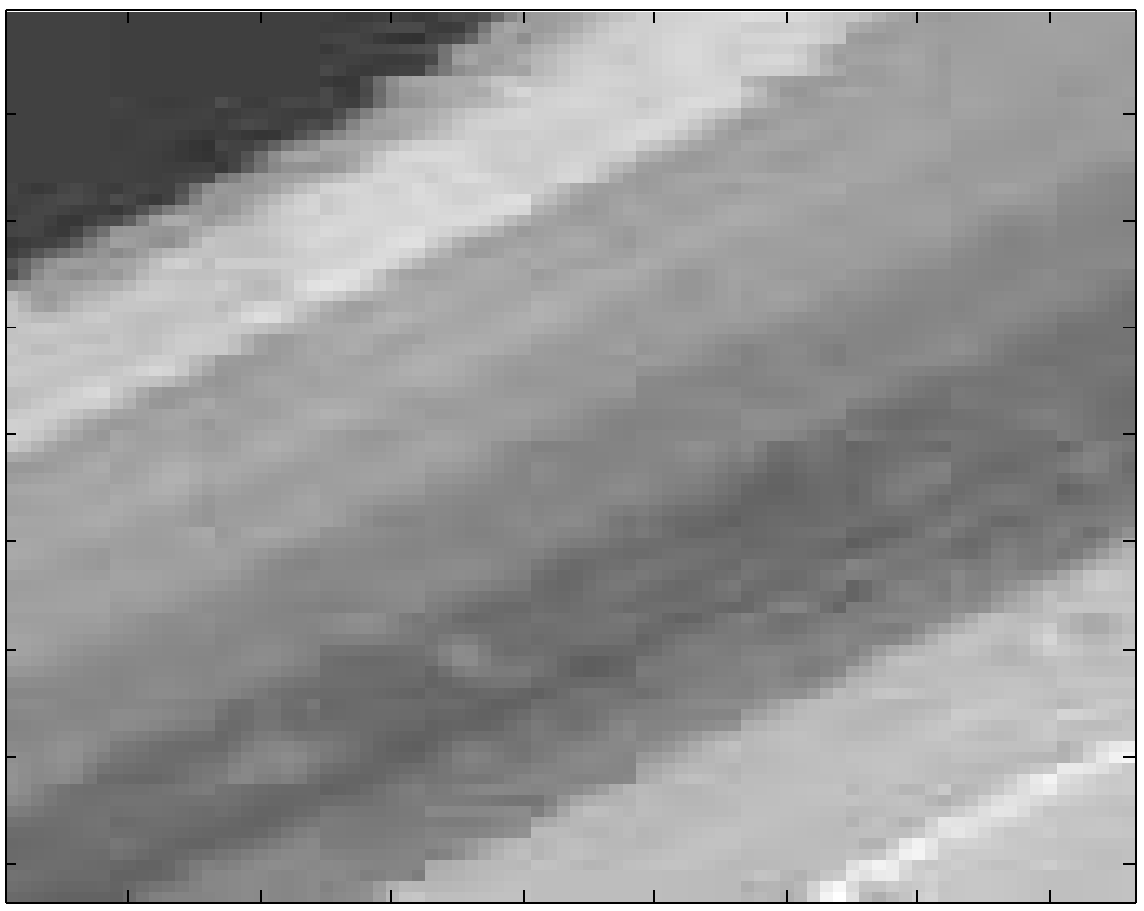}
}
 \subfigure[Ncut]{
\includegraphics[width=0.3\linewidth, height=0.2\linewidth]{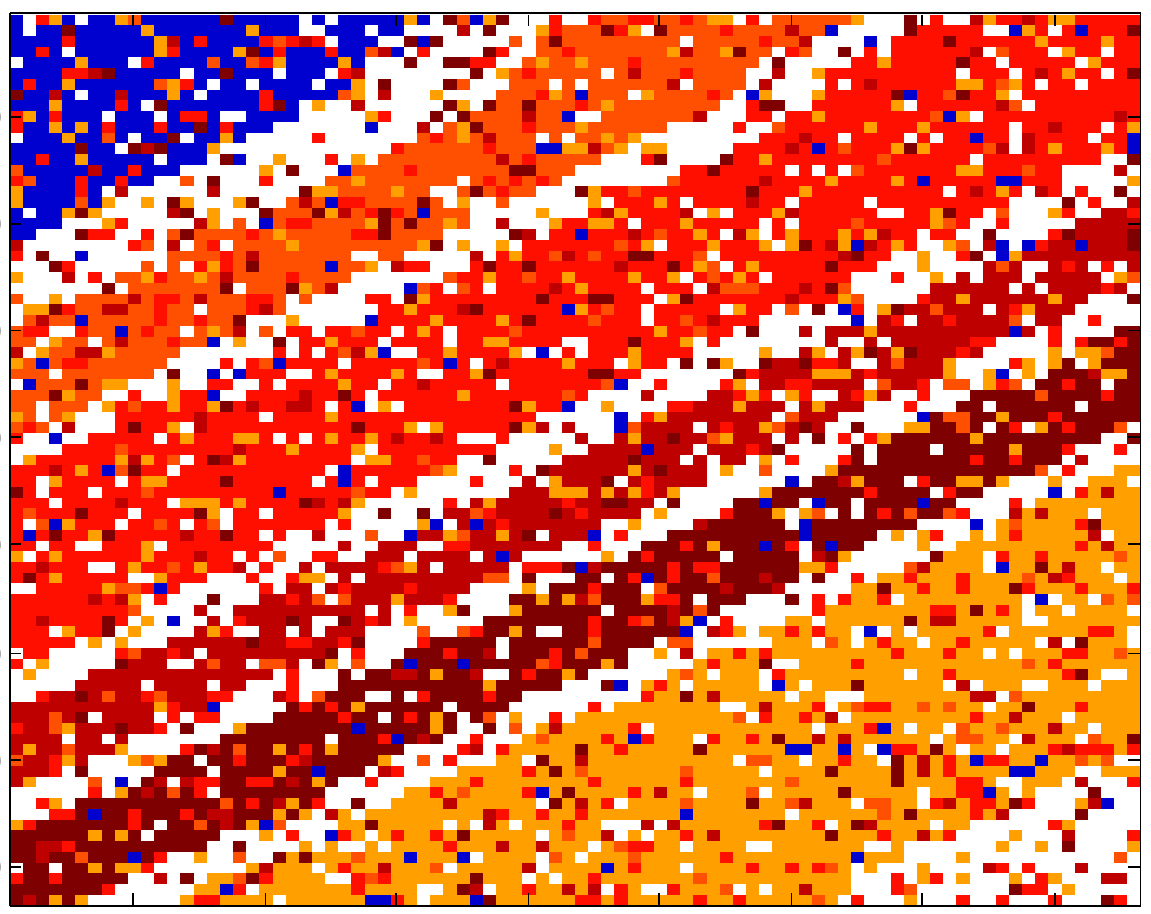}
}
 \subfigure[LRR+Ncut]{
\includegraphics[width=0.3\linewidth, height=0.2\linewidth]{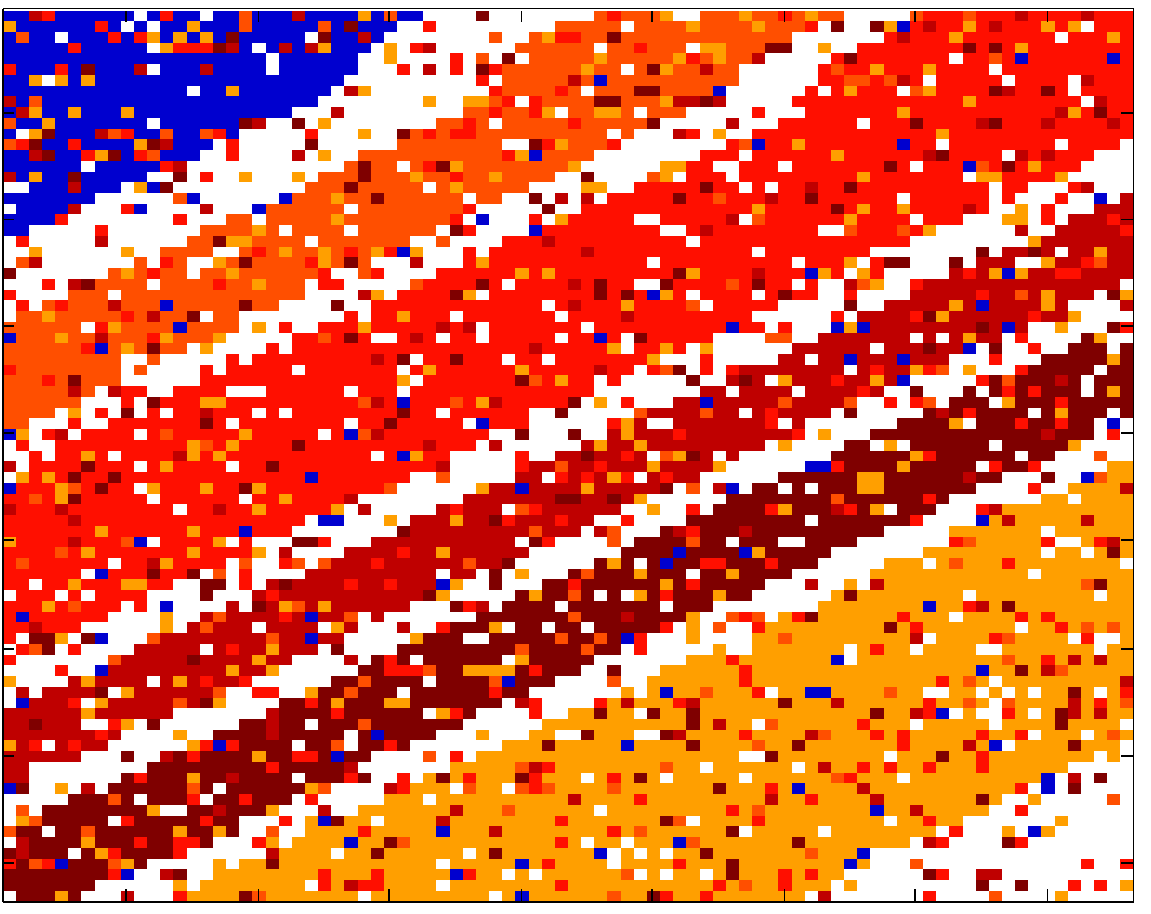}
}
 \subfigure[GNcut]{
\includegraphics[width=0.3\linewidth, height=0.2\linewidth]{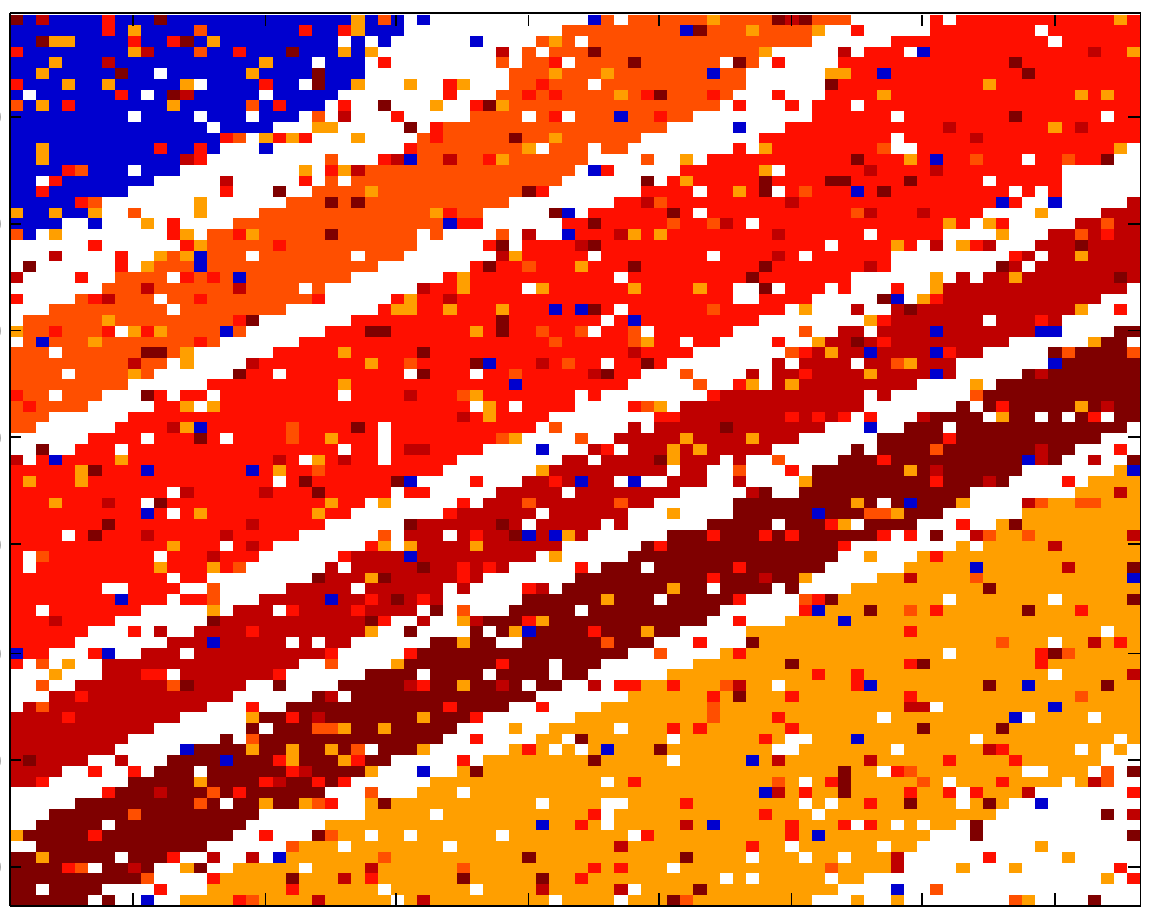}
}
 \subfigure[SC+Ncut]{
\includegraphics[width=0.3\linewidth, height=0.2\linewidth]{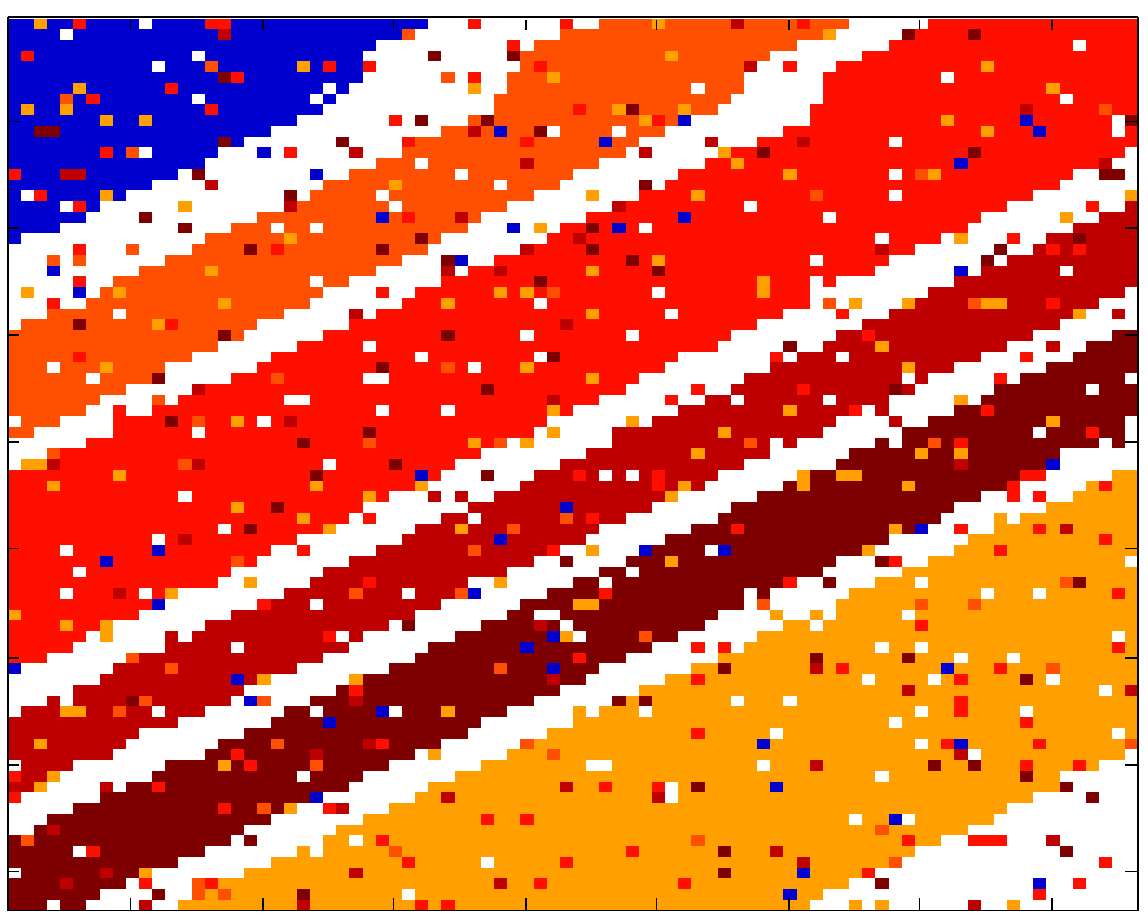}
}
 \subfigure[Proposed]{
\includegraphics[width=0.3\linewidth, height=0.2\linewidth]{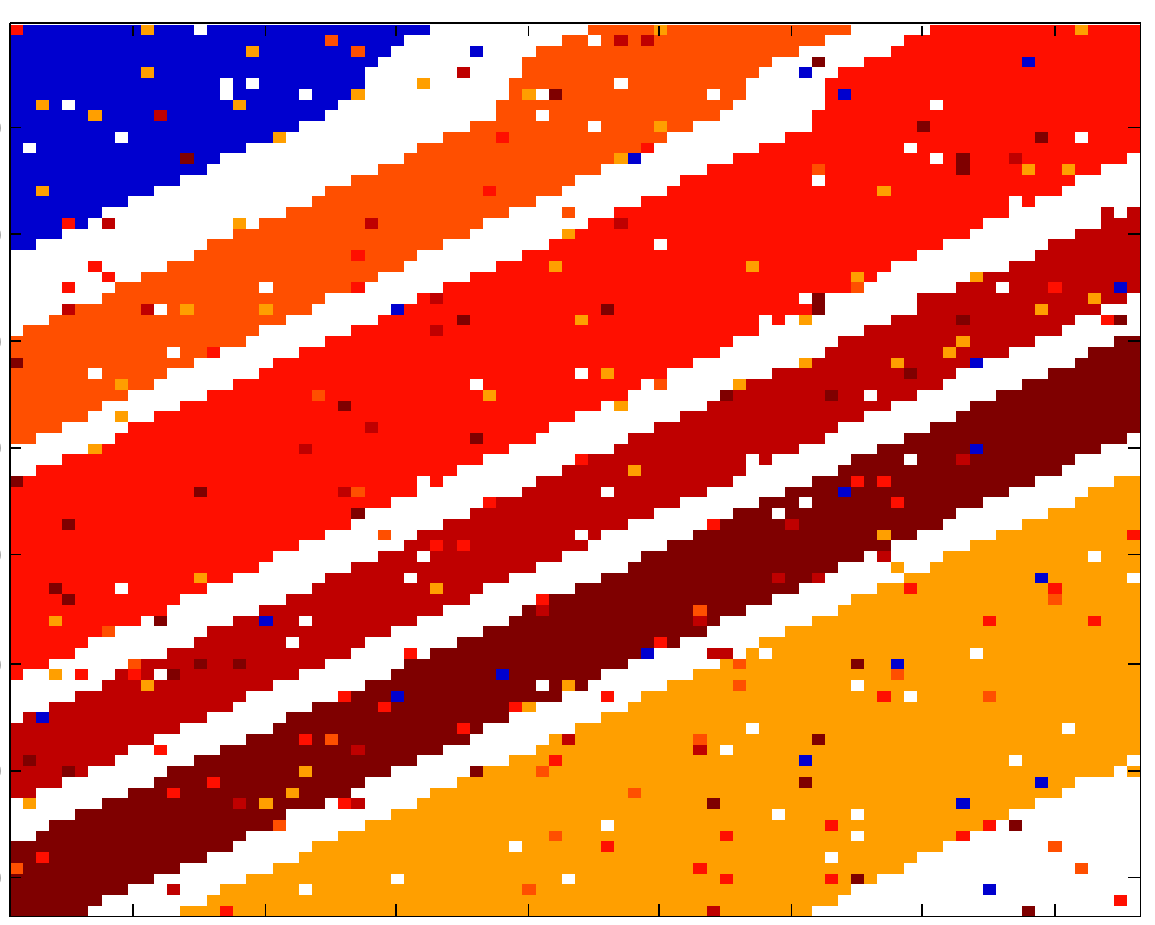}
}
\subfigure[Original Samson]{
\includegraphics[width=0.3\linewidth, height=0.2\linewidth]{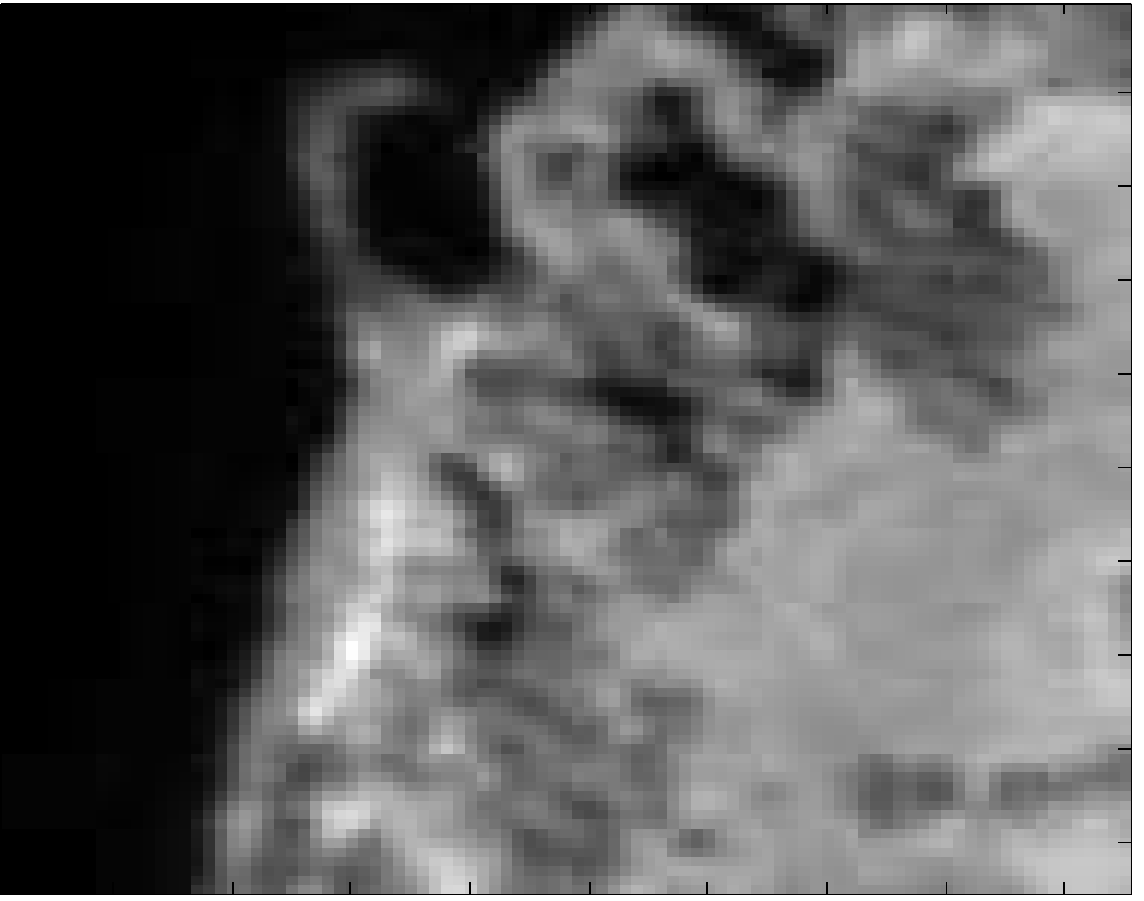}
}
 \subfigure[Ncut]{
\includegraphics[width=0.3\linewidth, height=0.2\linewidth]{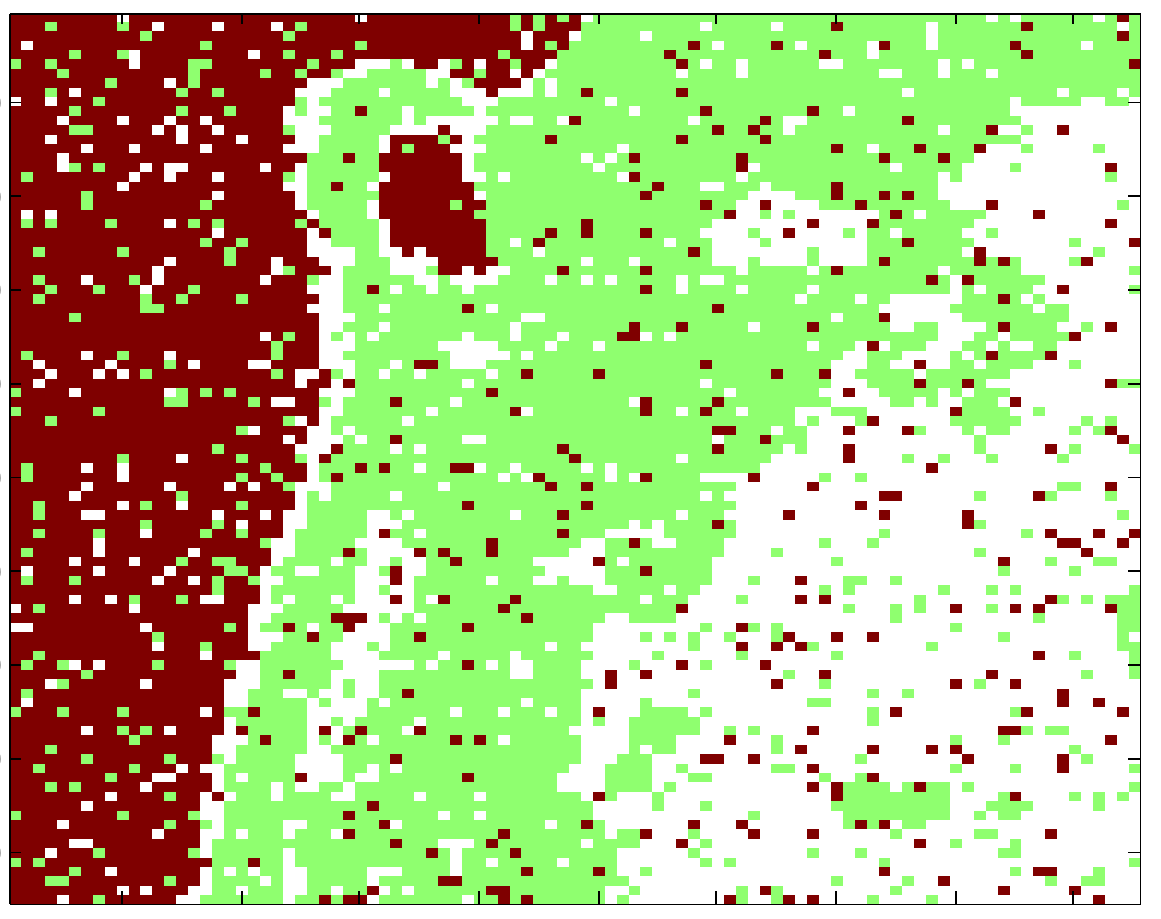}
}
 \subfigure[LRR+Ncut]{
\includegraphics[width=0.3\linewidth, height=0.2\linewidth]{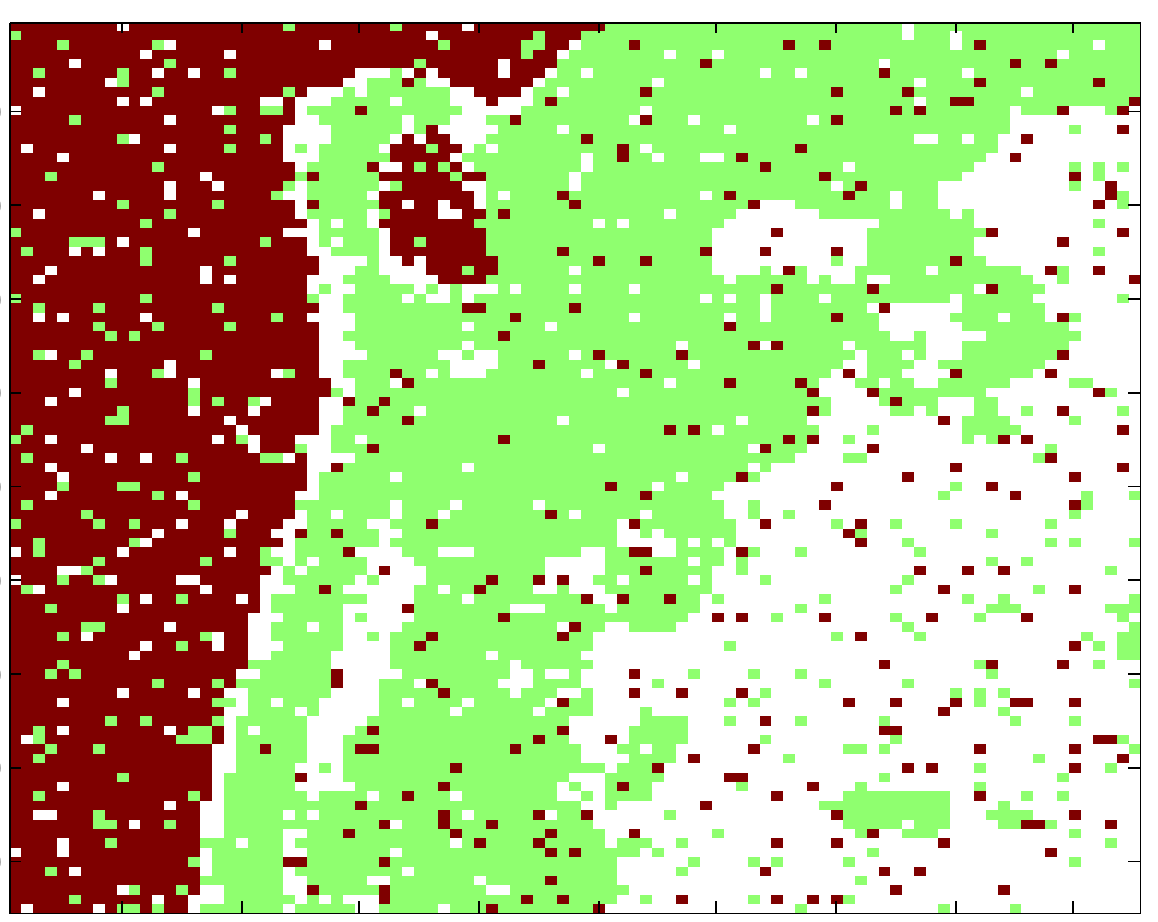}
}
 \subfigure[GNcut]{
\includegraphics[width=0.3\linewidth, height=0.2\linewidth]{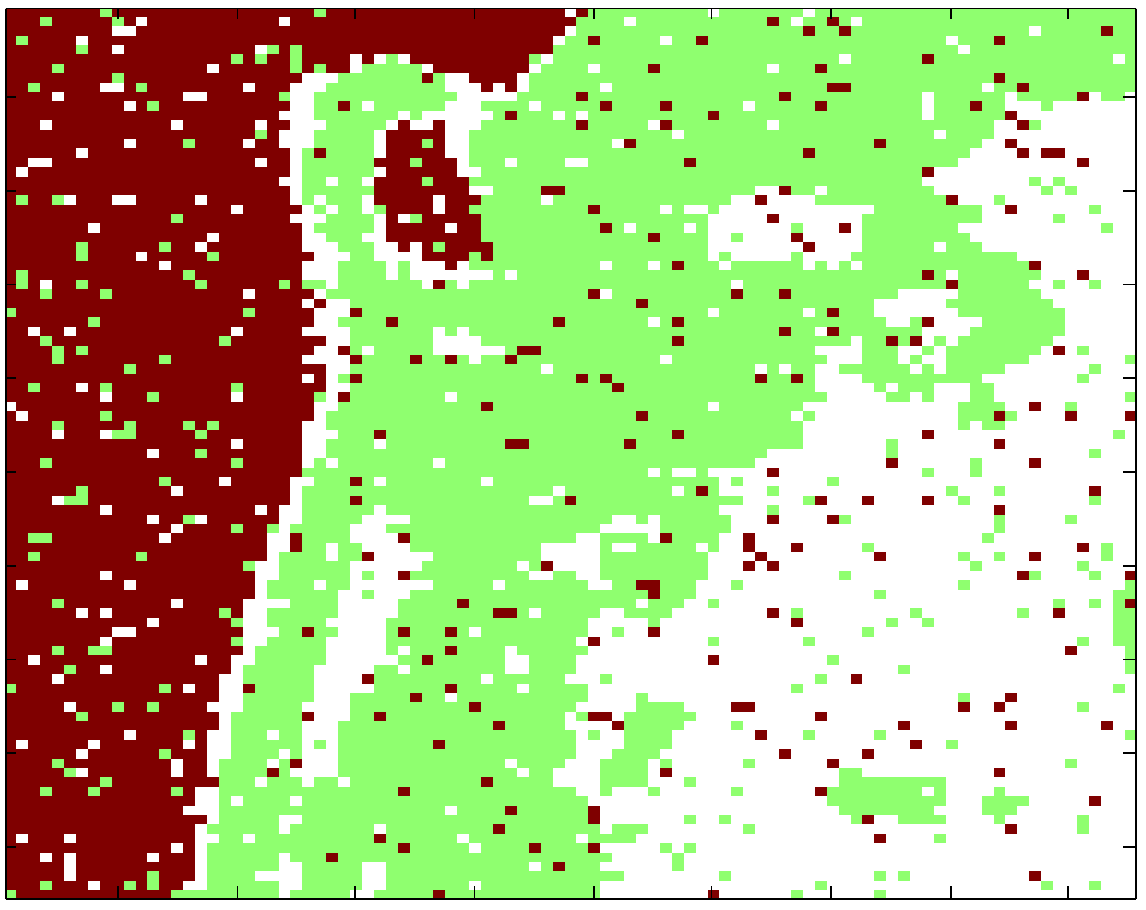}
}
 \subfigure[SC+Ncut]{
\includegraphics[width=0.3\linewidth, height=0.2\linewidth]{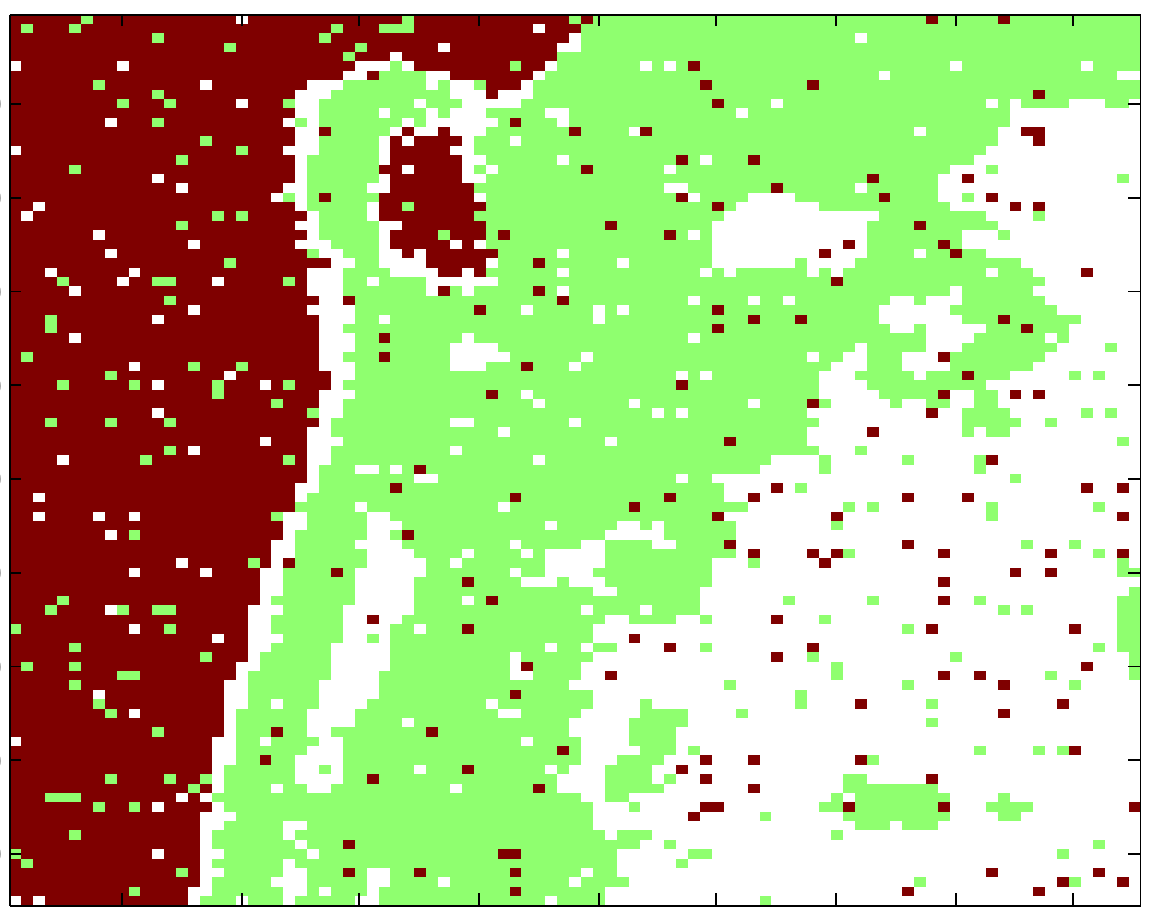}
}
 \subfigure[Proposed]{
\includegraphics[width=0.3\linewidth, height=0.2\linewidth]{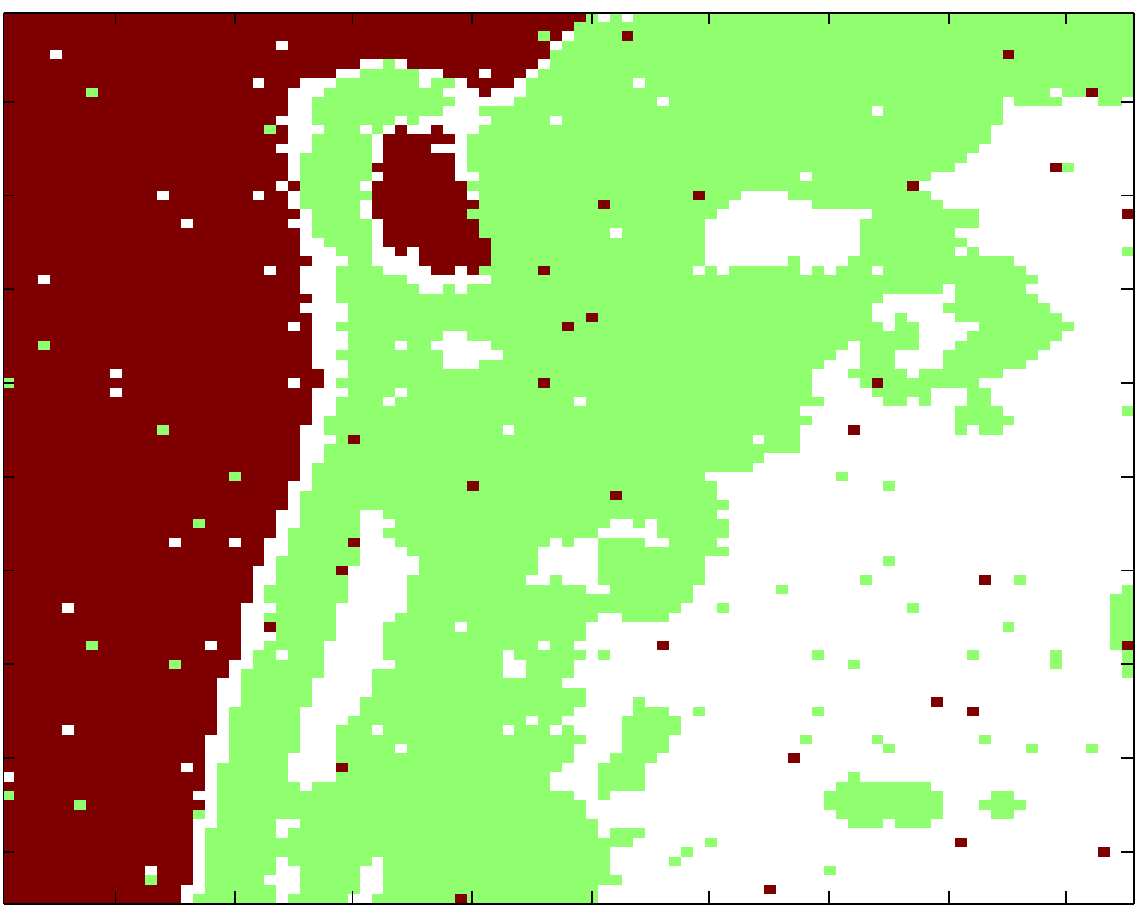}
}
\caption{Subspace Clustering results on PaviaCenter, Salinas-A and Samson}%
 \label{cluster123}
\end{figure*}
\addtocounter{figure}{-1}       
\begin{figure*}
\addtocounter{figure}{1}      
\centering
 \subfigure[Original Jasper Ridge]{
\includegraphics[width=0.3\linewidth, height=0.2\linewidth]{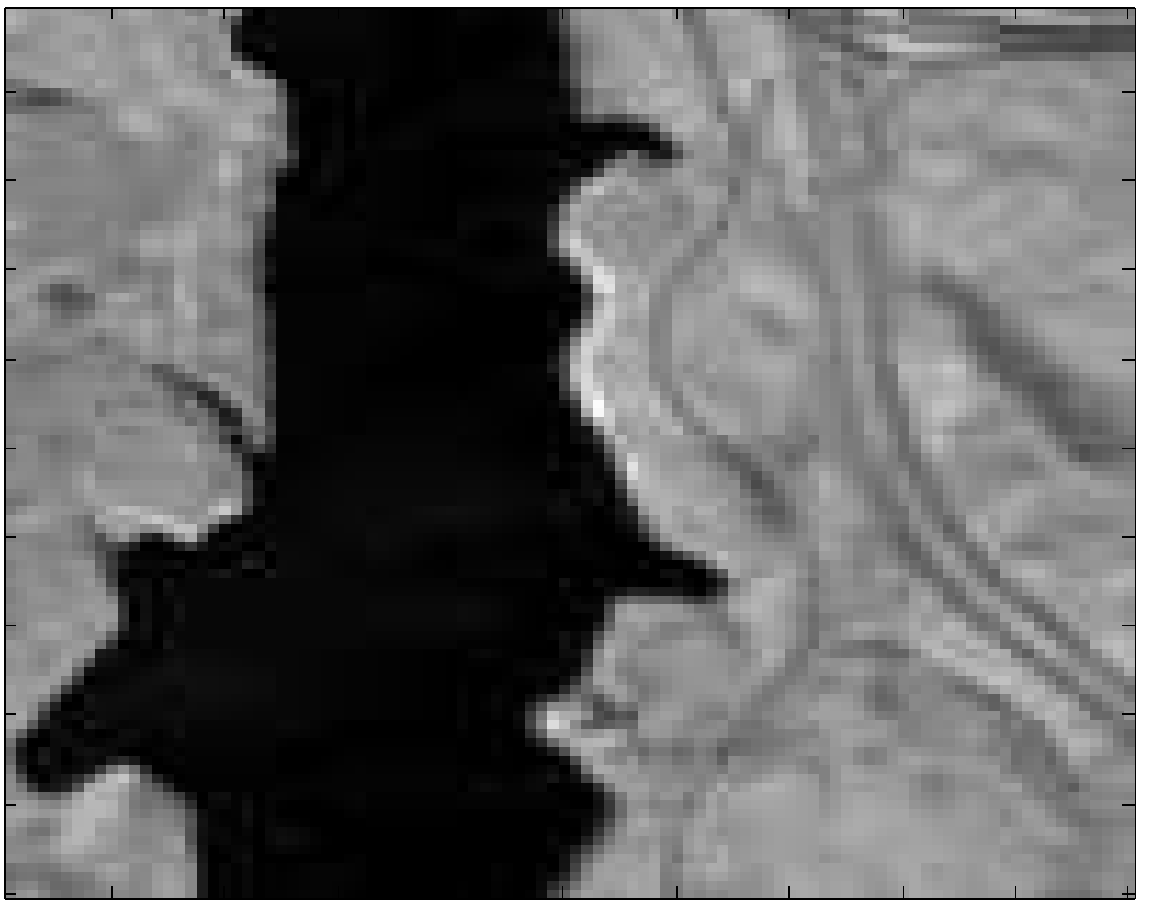}
}
 \subfigure[Ncut]{
\includegraphics[width=0.3\linewidth, height=0.2\linewidth]{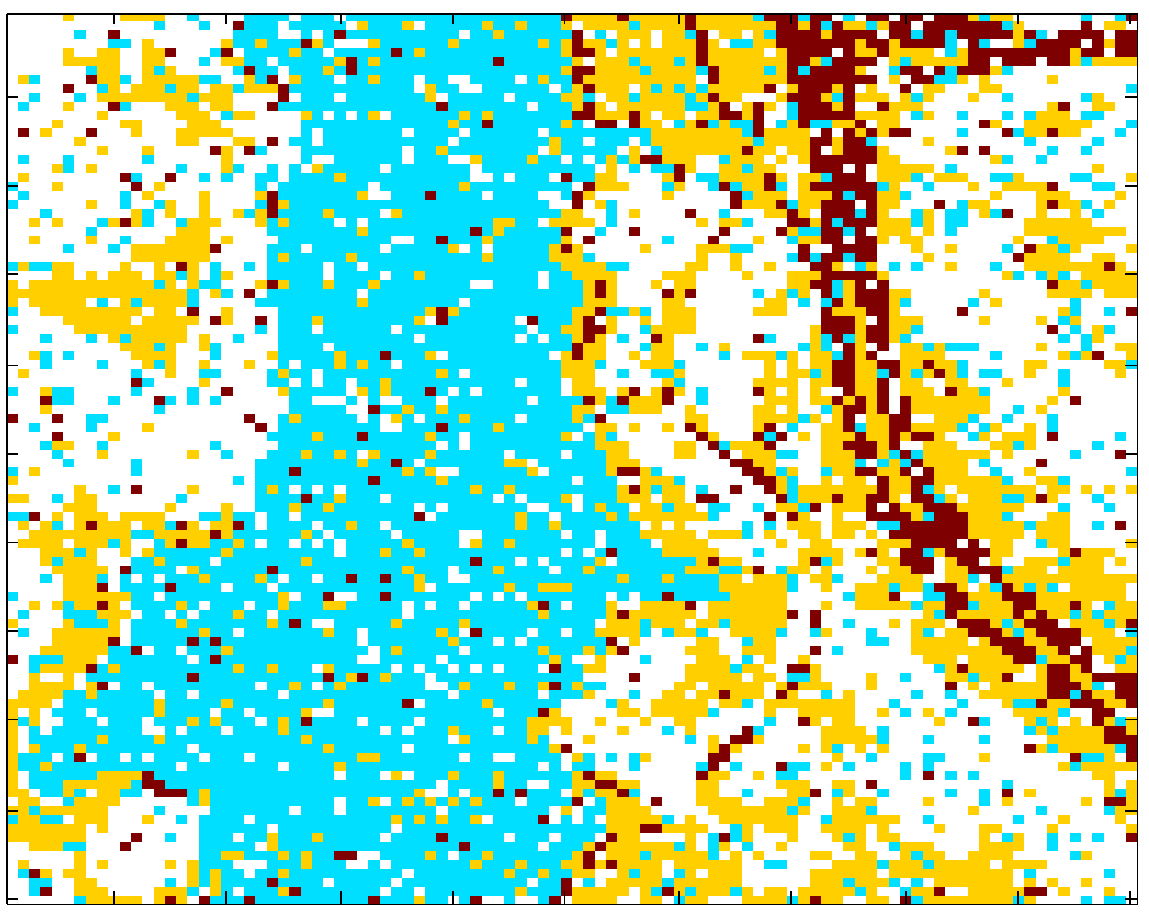}
}
 \subfigure[LRR+Ncut]{
\includegraphics[width=0.3\linewidth, height=0.2\linewidth]{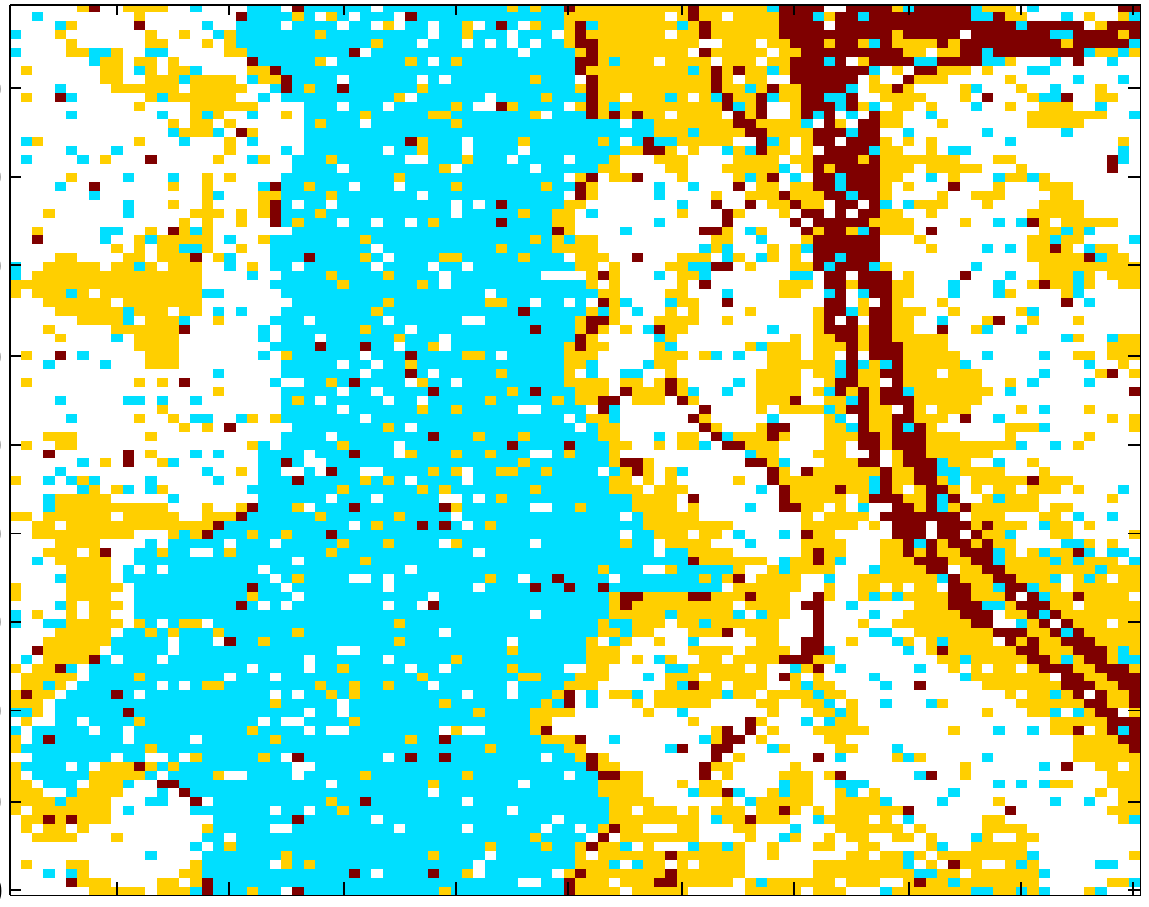}
}
 \subfigure[GNcut]{
\includegraphics[width=0.3\linewidth, height=0.2\linewidth]{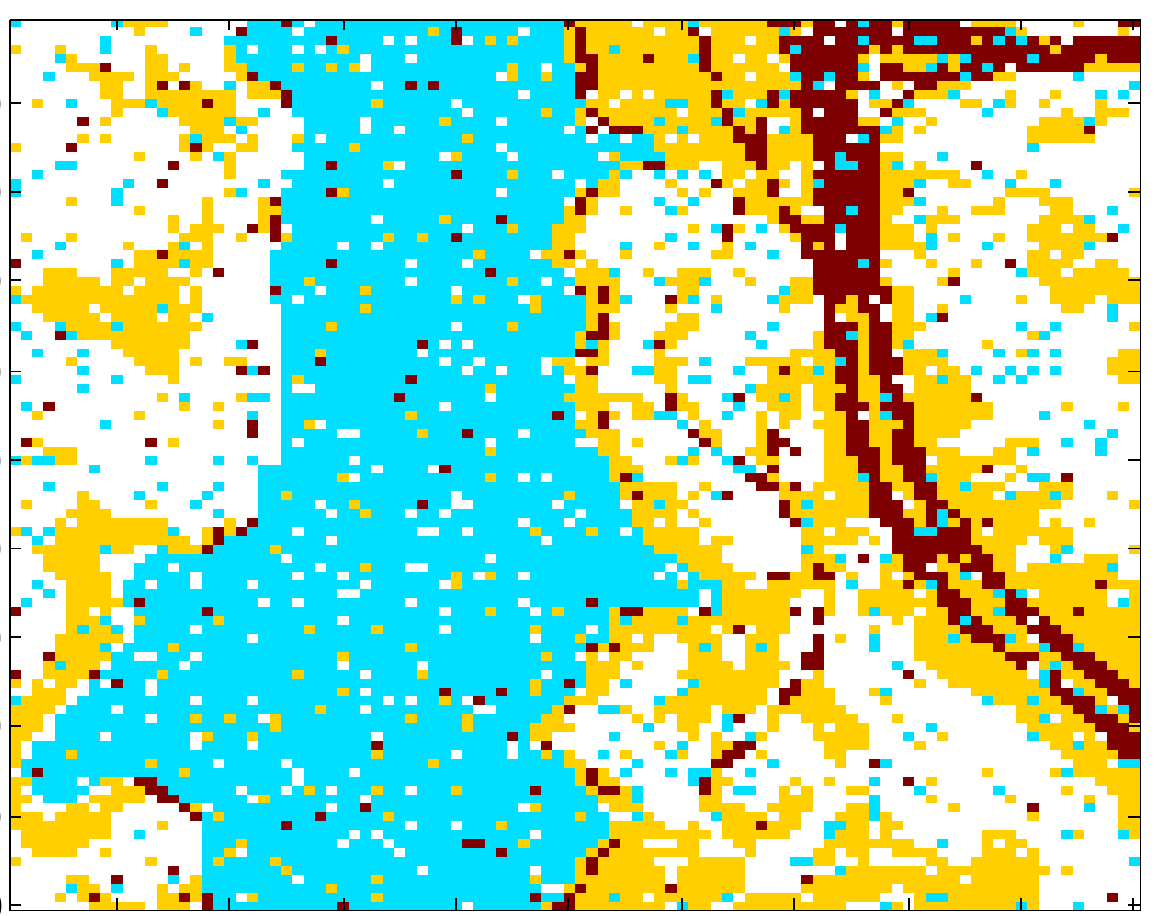}
}
 \subfigure[SC+Ncut]{
\includegraphics[width=0.3\linewidth, height=0.2\linewidth]{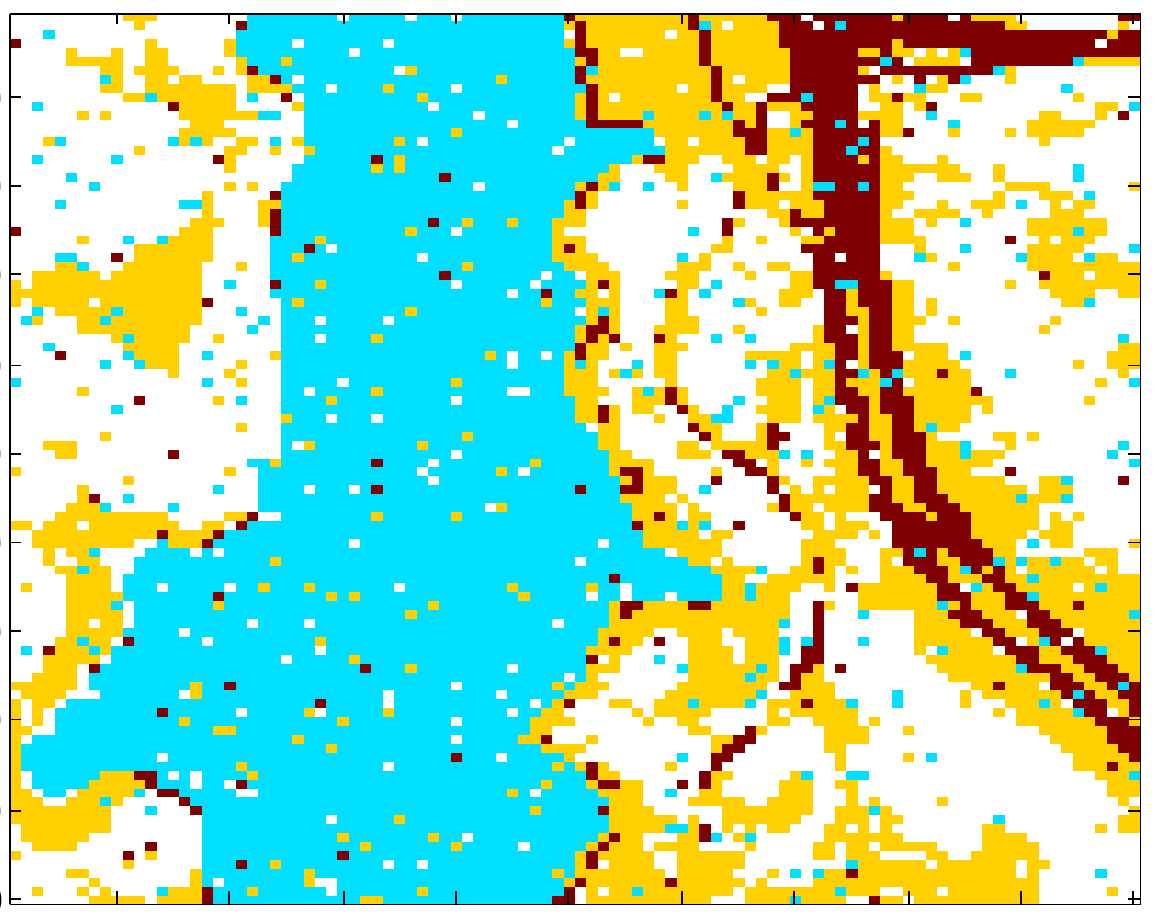}
}
 \subfigure[Proposed]{
\includegraphics[width=0.3\linewidth, height=0.2\linewidth]{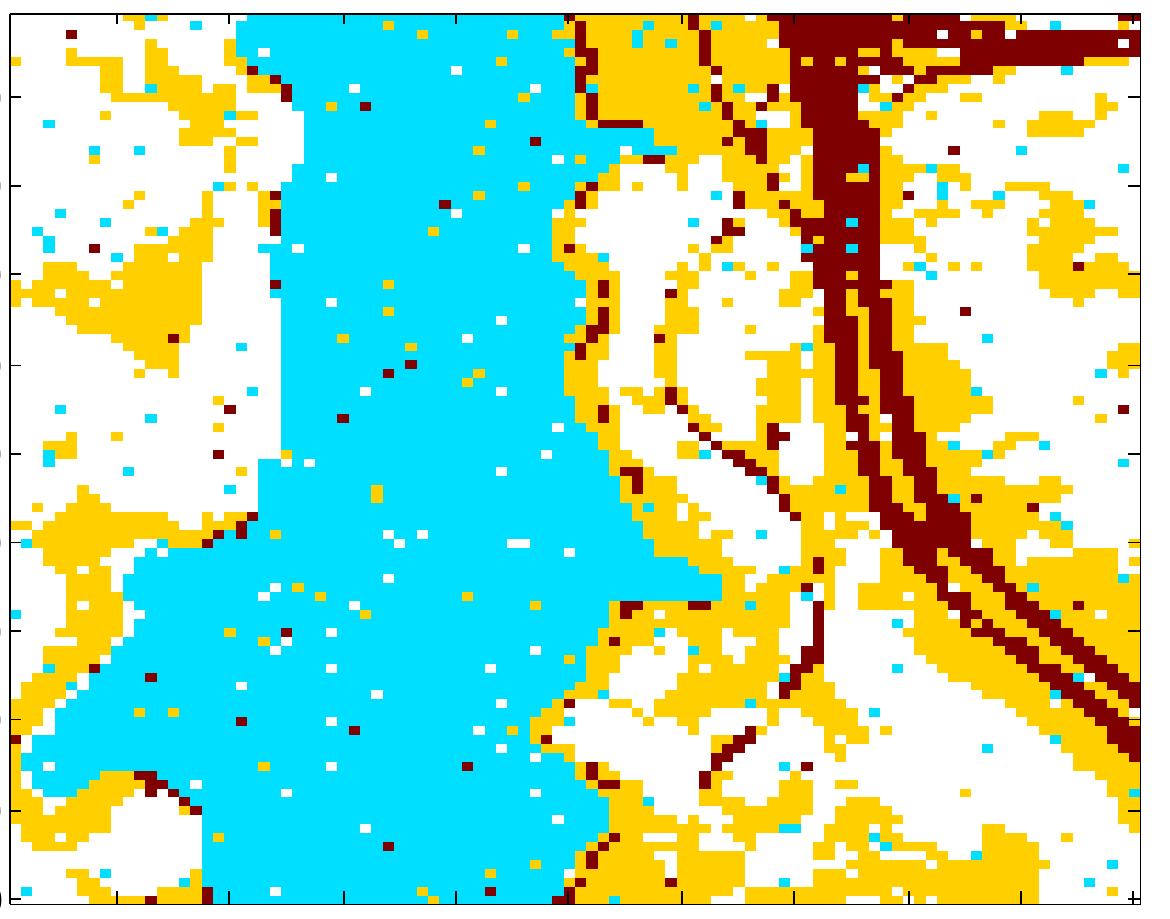}
}
\caption{Subspace Clustering results on Jasper Ridge}%
\label{cluster4}
\end{figure*}
\subsection{Performance on Corrupted Data}
\begin{figure*}

\centering
     \begin{minipage}[t]{0.33\linewidth}
       \subfigure[OASIS]{
       \label{fig:subfig:a} 
       \includegraphics[width=5.5cm,height=4cm]{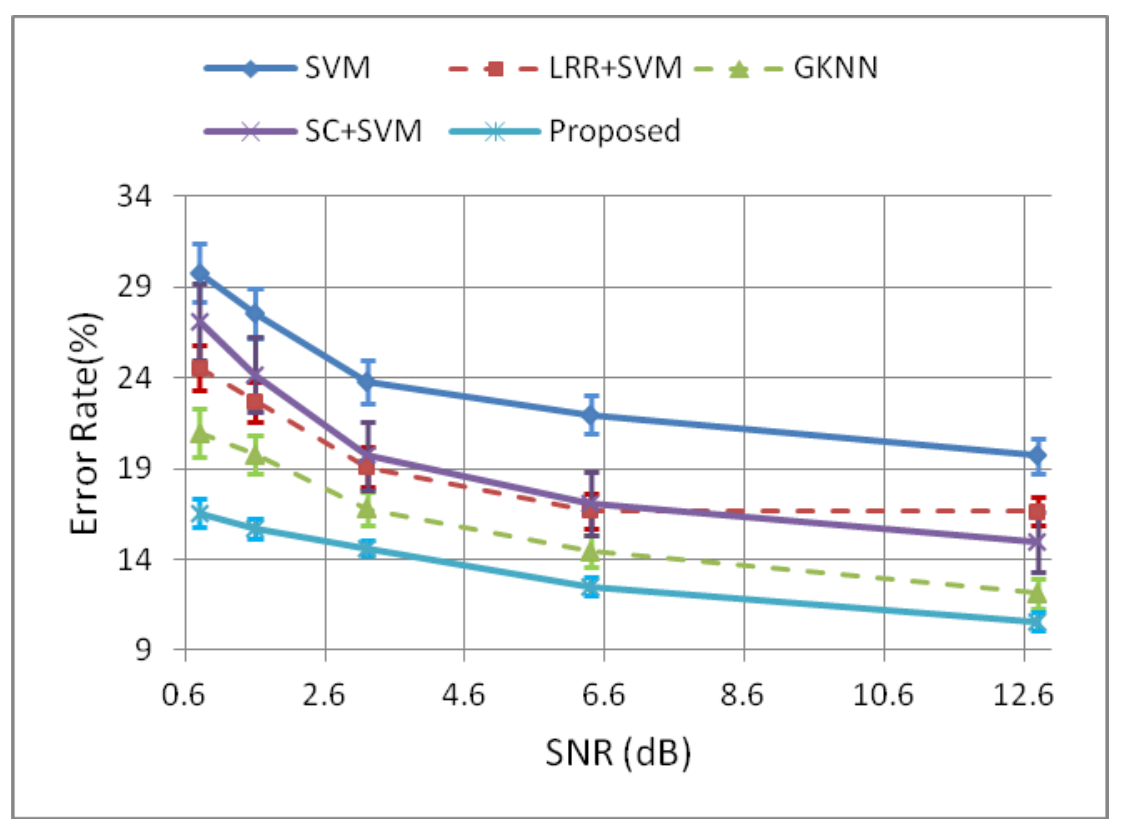}
     }
     \hspace{0.5in}
     \end{minipage}%
     \begin{minipage}[t]{0.33\linewidth}
     \subfigure[Lumbar]{
       \label{fig:subfig:b} 
       \includegraphics[width=5.5cm,height=4cm]{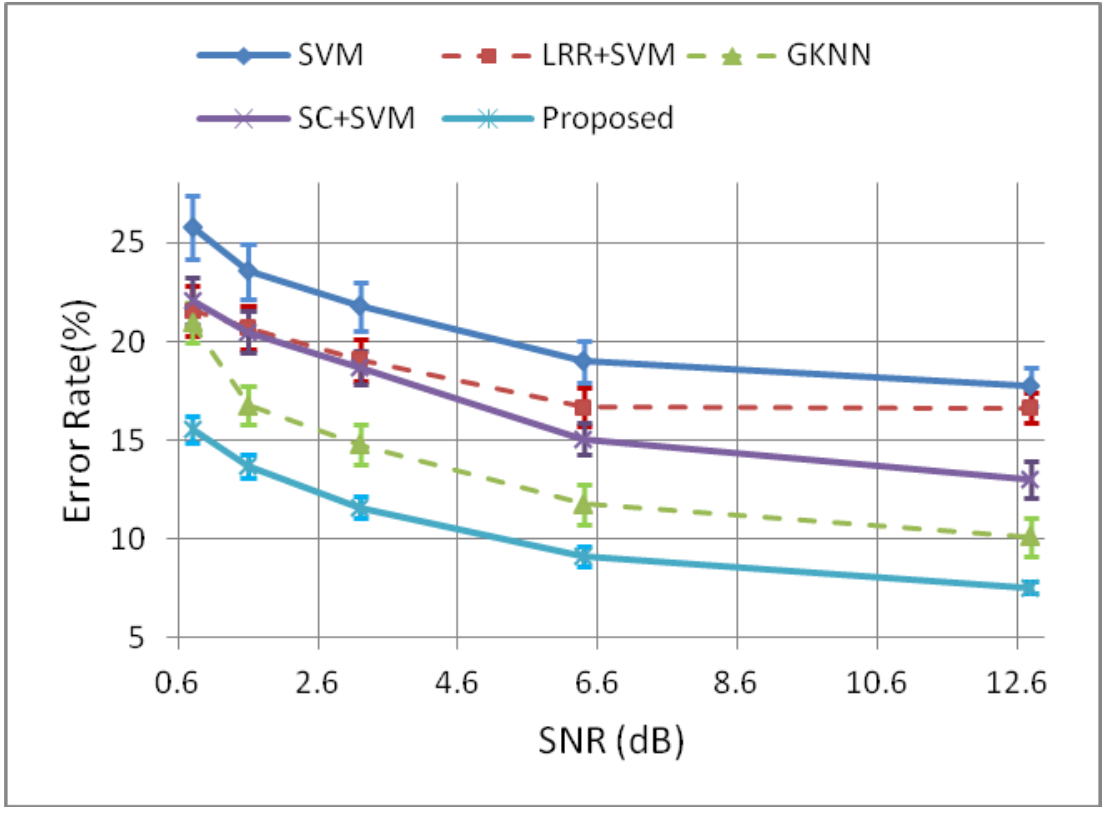}
     }
     \hspace{0.5in}
     \end{minipage}
     \begin{minipage}[t]{0.33\linewidth}
     \subfigure[Pavia Center]{
       \label{fig:subfig:c} 
       \includegraphics[width=5.5cm,height=4cm]{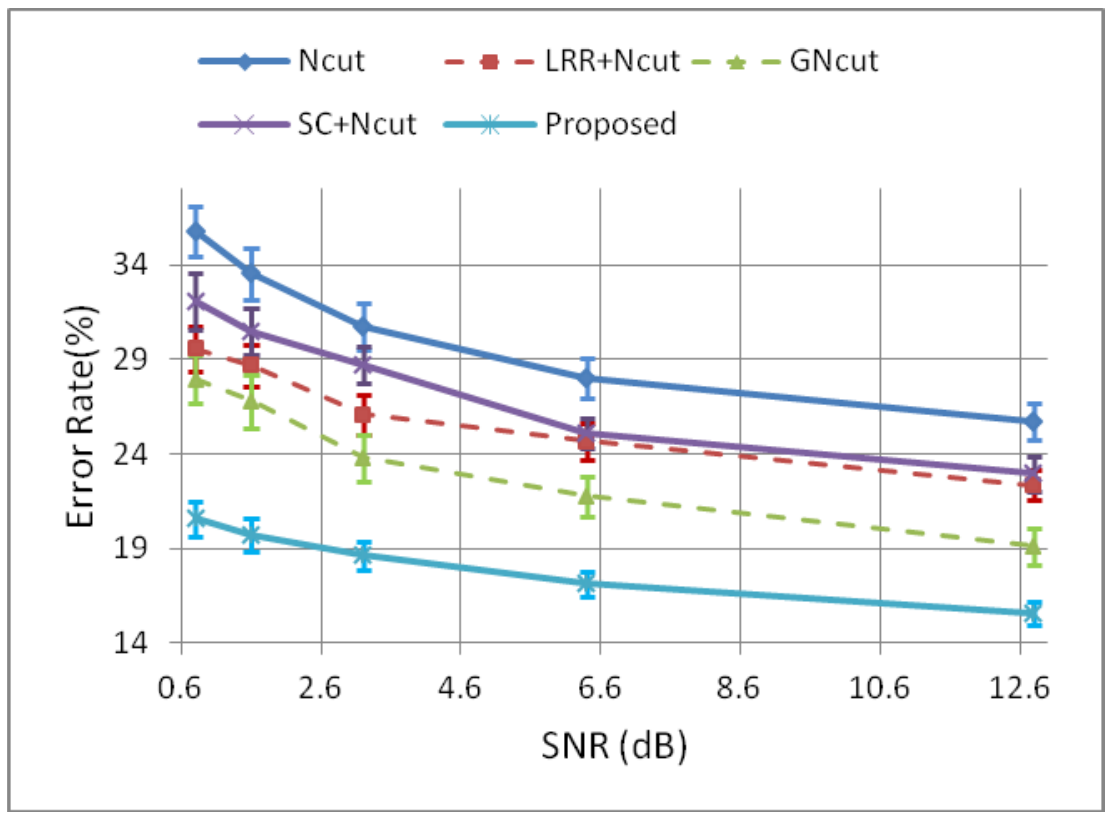}}
     \end{minipage}
       \begin{minipage}[t]{0.33\linewidth}
       \subfigure[Salinas-A]{
       \label{fig:subfig:a} 
       \includegraphics[width=5.5cm,height=4cm]{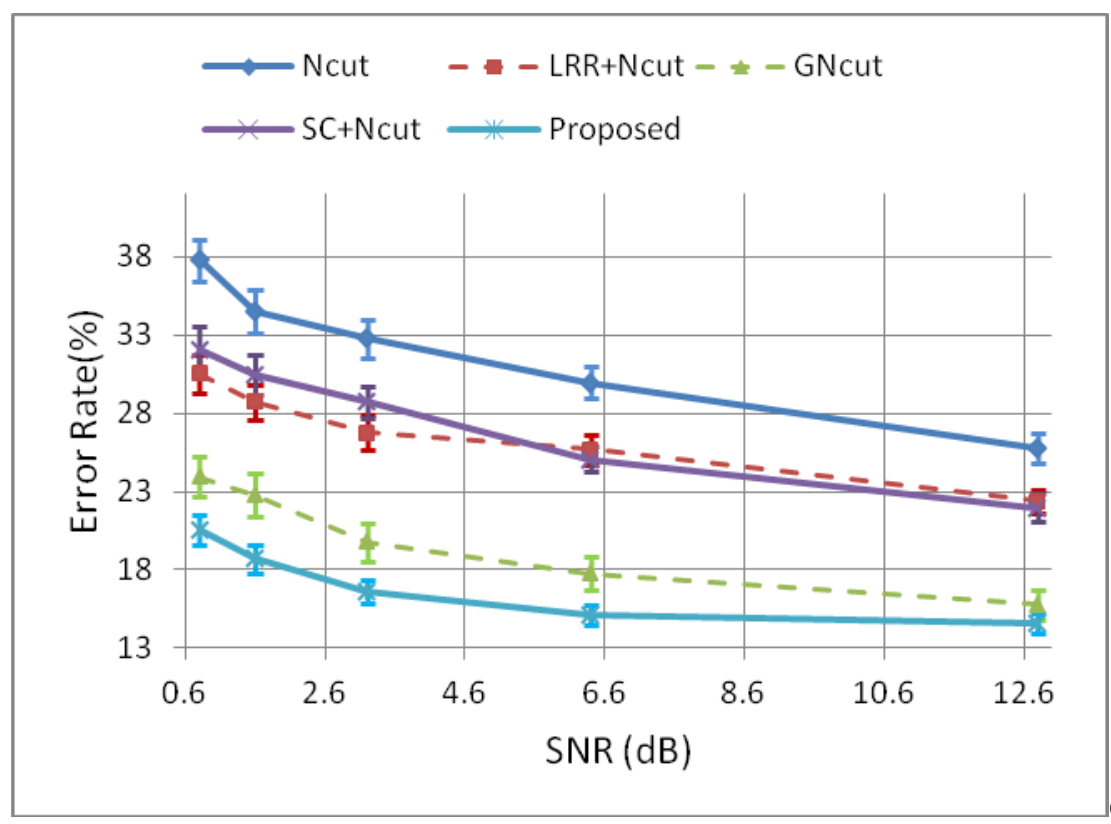}
     }
     \hspace{0.5in}
     \end{minipage}%
     \begin{minipage}[t]{0.33\linewidth}
     \subfigure[Samson]{
       \label{fig:subfig:b} 
       \includegraphics[width=5.5cm,height=4cm]{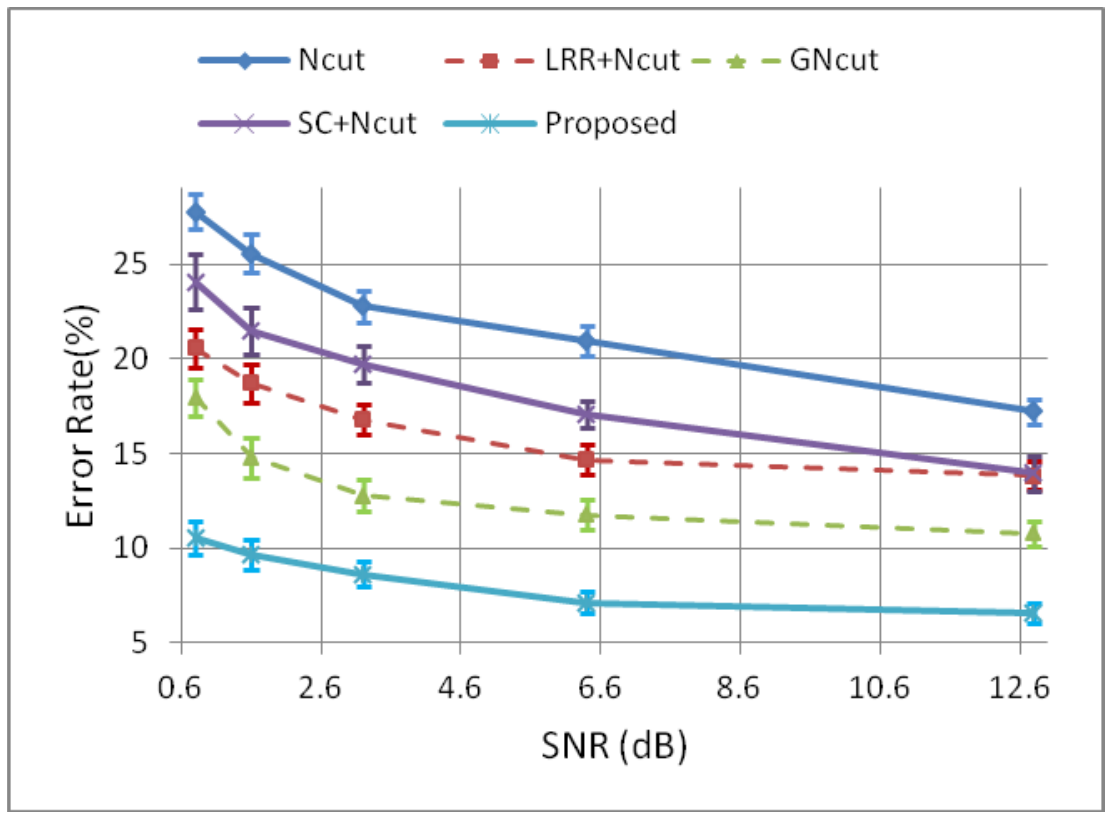}
     }
     \hspace{0.5in}
     \end{minipage}
     \begin{minipage}[t]{0.33\linewidth}
     \subfigure[Jasper Ridge]{
       \label{fig:subfig:c} 
       \includegraphics[width=5.5cm,height=4cm]{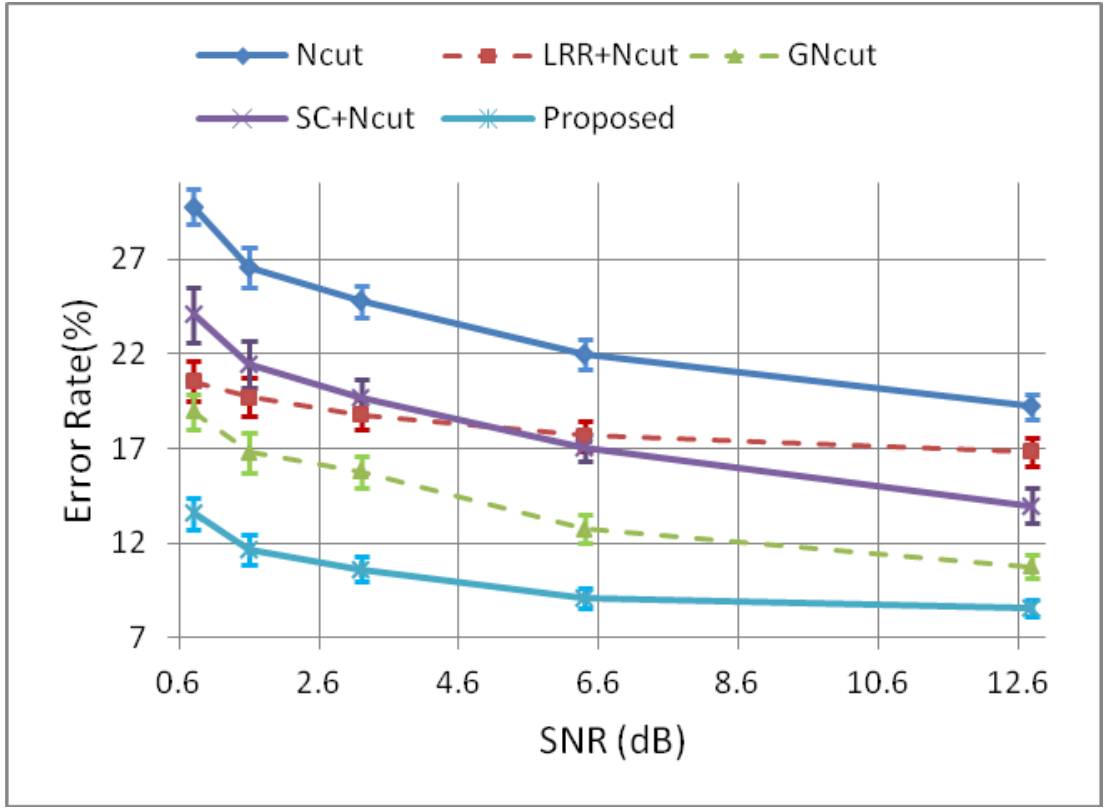}}
     \end{minipage}
\caption{Error rate comparisons with different SNR values on 6 datasets.} \label{fignoise}
\end{figure*}

 To further examine the noise robustness of the proposed model, we add Guassian white noises with different  signal-to-noise ratio (SNR) values to the above 6 baseline datasets used for both classification and subspace clustering tasks.  Figure \ref{fignoise} illustrates the performances on all the methods, with SNR values ranging from 0.8 to 12.8. Unlike SC related methods (SC+SVM and SC+Ncut) having comparable performances  on the clean datasets, their performances deteriorate when data are corrupted. In contrast, LRR based methods, no matter whether designed in the Euclidean space or on the manifold space,  always have stable performances with different SNR values,  which shows that LRR is more noisetolerable and robust than SC.
 Moreover, the geodesic distance based baselines like GKNN and GNcut outperform their linear Euclidean counterparts (i.e. LRR+SVM/Ncut, SVM/Ncut). This demonstrates that exploring data spatial correlation information in the intrinsic Riemannian geometry helps to boost prediction performance. We also observe that our new method performs best among all
  the methods on the corrupted data sets with different SNR values, confirming the importance of intrinsic geometry and sparse feature transform again.
\section{Conclusions}
In this paper, we propose a novel LRR model on the manifold of square root densities, in which we exploit the intrinsic property of the square root densities manifold in the Riemannian geometric context. Compared with the existing Euclidean LRR algorithms, the loss of the global linear structure is compensated by the local linear structures given by the tangent spaces of the manifold. A weight term  for sparse representation is also integrated into the model  to avoid assigning large weights to the far-away points on the manifold.  Furthermore, we derive an easily solvable optimization problem, which incorporates the structured embedding mapping and  the intrinsic geodesic distance on the manifold  into the LRR model. Our experiments demonstrate that our proposed method is efficient and robust to  noise, and produces superior results compared to other state-of-the-art methods for classification and subspace clustering applications on computer vision datasets.

%
\bibliographystyle{abbrv}

\begin{thebibliography}{10}

\bibitem{DICOM}
Dicom sample image sets.
\newblock \url{http://www.osirix-viewer.com/datasets/}.

\bibitem{ArsignyFillardPennecAyache2006}
V.~Arsigny, P.~Fillard, X.~Pennec, and N.~Ayache.
\newblock {Log-Euclidean} metrics for fast and simple calculus on diffusion
  tensors.
\newblock In {\em Magnetic Resonance in Medicine}, 2006.

\bibitem{BaoCaiJi2013}
C.~Bao, J.~Cai, and H.~Ji.
\newblock Fast sparsity-based orthogonal dictionary learning for image
  restoration.
\newblock In {\em IEEE International Conference on Computer Vision (ICCV)},
  pages 3384--3391, Dec 2013.

\bibitem{CaiCandasShen2010}
J.~Cai, E.~J. Candas, and Z.~Shen.
\newblock A singular value thresholding algorithm for matrix completion.
\newblock {\em SIAM Journal on Optimization}, 20(4):1956--1982, 2010.

\bibitem{Callaghan1994}
P.~Callaghan.
\newblock {\em Principles of Nuclear Magnetic Resonance Microscopy}.
\newblock Oxford Science Publications. Oxford University Press, USA, 1994.

\bibitem{ChangLin2001}
C.-c. Chang and C.-J. Lin.
\newblock Libsvm: a library for support vector machines, 2001.

\bibitem{ChenWangLu1992}
G.-H. Chen, B.~Wang, and C.~Lu.
\newblock On the parallel computation of the algebraic path problem.
\newblock {\em IEEE Transactions on Parallel and Distributed Systems},
  3(2):251--256, Mar 1992.

\bibitem{EldarNeedellPlan2012}
Y.~C. Eldar, D.~Needell, and Y.~Plan.
\newblock Uniqueness conditions for low-rank matrix recovery.
\newblock {\em Applied and Computational Harmonic Analysis}, 33(2):309 -- 314,
  2012.

\bibitem{ElhamifarVidal2013}
E.~Elhamifar and R.~Vidal.
\newblock Sparse subspace clustering: Algorithm, theory, and applications.
\newblock {\em IEEE Trans. Pattern Anal. Mach. Intell.}, 35(11):2765--2781,
  Nov. 2013.

\bibitem{FauvelBenediktssonChanussotSveinsson2008}
M.~Fauvel, J.~Benediktsson, J.~Chanussot, and J.~Sveinsson.
\newblock Spectral and spatial classification of hyperspectral data using
  {SVMs} and morphological profiles.
\newblock {\em IEEE Transactions on Geoscience and Remote Sensing},
  46:3804--3814, 2008.

\bibitem{FuGaoHongTien2015}
Y.~Fu, J.~Gao, X.~Hong, and D.~Tien.
\newblock Low rank representation on {Riemannian} manifold of symmetric
  positive definite matrices.
\newblock In {\em SIAM International Conference on Data Mining (SDM)}, 2015.

\bibitem{Jones2011}
D.~K. Jones.
\newblock {Diffusion MRI theory, methods, and applications}.
\newblock \url{http://www.worldcat.org/isbn/9780195369779}, 2011.

\bibitem{KongWang2012}
S.~Kong and D.~Wang.
\newblock A dictionary learning approach for classification: Separating the
  particularity and the commonality.
\newblock In {\em European Conference on Computer Vision (ECCV)}, volume 7572
  of {\em Lecture Notes in Computer Science}, pages 186--199. Springer Berlin
  Heidelberg, 2012.

\bibitem{Lee2003}
J.~M. Lee.
\newblock {\em Introduction to smooth manifolds}.
\newblock Graduate texts in mathematics. Springer, New York, Berlin,
  Heidelberg, 2003.

\bibitem{LinChenMa2010}
Z.~Lin, M.~Chen, and Y.~Ma.
\newblock {The Augmented Lagrange Multiplier Method for Exact Recovery of
  Corrupted Low-Rank Matrices}.
\newblock {\em Mathematical Programming}, 2010.

\bibitem{LinLiuSu2011}
Z.~Lin, R.~Liu, and Z.~Su.
\newblock Linearized alternating direction method with adaptive penalty for
  low-rank representation.
\newblock In {\em Advances in Neural Information Processing Systems}, pages
  612--620. Curran Associates, Inc., 2011.

\bibitem{LiuLinYanSunYuMa2013}
G.~Liu, Z.~Lin, S.~Yan, J.~Sun, Y.~Yu, and Y.~Ma.
\newblock Robust recovery of subspace structures by low-rank representation.
\newblock {\em IEEE Transactions on Pattern Analysis and Machine Intelligence},
  35(1):171--184, 2013.

\bibitem{LiuLiuYu2010}
G.~Liu, Z.~Liu, and Y.~Yu.
\newblock Robust subspace segmentation by low-rank representation.
\newblock In {\em International Conference on Machine Learning (ICML)}, 2010.

\bibitem{LiuYangGaoYinChen2014}
H.~Liu, M.~Yang, Y.~Gao, Y.~Yin, and L.~L. Chen.
\newblock Bilinear discriminative dictionary learning for face recognition.
\newblock {\em Pattern Recognition}, 47(5):1835 -- 1845, 2014.

\bibitem{liuchezhangxu2014}
J.~Liu, Y.~Chen, J.~Zhang, and Z.~Xu.
\newblock Enhancing low-rank subspace clustering by manifold regularization.
\newblock {\em IEEE Transaction on Image Processing}, 23(9):4022--4030, 2014.

\bibitem{MaSuJurie2012}
B.~Ma, Y.~Su, and F.~Jurie.
\newblock {BiCov: a novel image representation for person re-identification and
  face verification}.
\newblock In {\em {British Machive Vision Conference (BMVC)}}, 2012.

\bibitem{MarcusWangParkerCsernanskyMorrisBuckner2007}
D.~S. Marcus, T.~H. Wang, J.~Parker, J.~G. Csernansky, J.~C. Morris, and R.~L.
  Buckner.
\newblock Open access series of imaging studies ({OASIS}): Cross-sectional mri
  data in young, middle aged, nondemented, and demented older adults.
\newblock {\em Journal of Cognitive Neuroscience}, 19:1498--1507, 2007.

\bibitem{MyronenkoSong2010}
A.~Myronenko and X.~Song.
\newblock Intensity-based image registration by minimizing residual complexity.
\newblock {\em IEEE Transactions on Medical Imaging}, 29(11):1882--1891, Nov
  2010.

\bibitem{PangYuanLi2008}
Y.~Pang, Y.~Yuan, and X.~Li.
\newblock Gabor-based region covariance matrices for face recognition.
\newblock {\em IEEE Transactions on Circuits and Systems for Video Technology},
  18(7):989--993, July 2008.

\bibitem{PlazaMartinezPerezPlaza2004}
P.~Plaza, Antonio J.and~Martinez, R.~M. Perez, and J.~Plaza.
\newblock Hyperspectral image analysis by scale-orientation morphological
  profiles.
\newblock {\em Image and signal processing for remote sensing}, 9:432--439,
  2004.

\bibitem{shalitweinshallchechik2012}
U.~Shalit, D.~Weinshall, and G.~Chechik.
\newblock Online learning in the embedded manifold of low-rank matrices.
\newblock {\em Journal of Machine Learning Research}, 13:429--458, 2012.

\bibitem{ShenWenZhang2014}
Y.~Shen, Z.~Wen, and Y.~Zhang.
\newblock Augmented {Lagrangian} alternating direction method for matrix
  separation based on low-rank factorization.
\newblock {\em Optimization Methods Software}, 29(2):239--263, Mar. 2014.

\bibitem{ShiMalik2000}
J.~Shi and J.~Malik.
\newblock Normalized cuts and image segmentation.
\newblock {\em IEEE Transaction on Pattern Analysis and Machine Intelligence},
  22(8):888--905, Aug. 2000.

\bibitem{SrivastavaJermynJoshi2007}
A.~Srivastava, I.~Jermyn, and S.~Joshi.
\newblock Riemannian analysis of probability density functions with
  applications in vision.
\newblock In {\em IEEE Conference on Computer Vision and Pattern Recognition
  (CVPR )}, pages 1--8, June 2007.

\bibitem{SunXieYeHoEntezariBlackbankVemuri2013}
J.~Sun, Y.~Xie, W.~Ye, J.~Ho, A.~Entezari, S.~J. Blackband, and B.~C. Vemuri.
\newblock Dictionary learning on the manifold of square root densities and
  application to reconstruction of diffusion propogator fields.
\newblock {\em Inf Process Med Imaging}, 23:619--631, 2013.

\bibitem{vandereyckenasilvandewalle2013}
B.~Vandereycken, P.~A. Asil, and S.~Vandewalle.
\newblock A {Riemannian} geometry with complete geodesics for the set of
  positive semidefinite matrices of fixed rank.
\newblock {\em IMA Journal of Numerical Analysis}, 33:481--514, 2013.

\bibitem{wanghugaosunyin2014}
B.~Wang, Y.~Hu, J.~Gao, Y.~Sun, and B.~Yin.
\newblock Low rank representation on {Grassmann} manifold.
\newblock In {\em Asian Conference on Computer Vision (ACCV)}, 2014.

\bibitem{XieHoVemuri2013}
Y.~Xie, J.~Ho, and B.~Vemuri.
\newblock On a nonlinear generalization of sparse coding and dictionary
  learning.
\newblock In {\em International Confernenceon Machine Learning (ICML)}, 2013.

\bibitem{zhangzhao2013}
Z.~Zhang and K.~Zhao.
\newblock Low-rank matrix approximation with manifold regularization.
\newblock {\em IEEE Transaction on Pattern Analysis and Machine Intelligence},
  35(7):1717--1729, 2013.

\bibitem{ZhuWangXiangFanPan2014}
F.~Zhu, Y.~Wang, S.~Xiang, B.~Fan, and C.~Pan.
\newblock Structured sparse method for hyperspectral unmixing.
\newblock {\em ISPRS Journal of Photogrammetry and Remote Sensing},
  88:101--118, 2014.

\end{thebibliography}

\end{document}